\documentclass[preprint,11pt,authoryear]{elsarticle}

\usepackage[a4paper,margin=2.3cm]{geometry}
\usepackage{setspace}
\usepackage{enumitem}
\setlist{itemsep=0.2em,topsep=0.3em,parsep=0pt,partopsep=0pt}

\usepackage{amssymb}
\usepackage{lipsum}
\usepackage{xcolor}
\definecolor{RevisionText}{HTML}{000000}
\newcommand{\rev}[1]{{\color{RevisionText}#1}}

\usepackage{needspace}
\newsavebox{\comparisoncaptionbox}
\newlength{\comparisonheight}

\usepackage[labelfont=bf,labelsep=period]{caption}
\usepackage{amsmath}
\usepackage{booktabs}
\usepackage{multirow}
\usepackage{array}
\newcommand{\tableformat}{\fontsize{10}{12}\selectfont\setlength{\tabcolsep}{2.5pt}\renewcommand{\arraystretch}{1.15}}
\newcommand{\metriccell}[2]{$#1\pm#2$}
\newcommand{\bestmetriccell}[2]{$\boldsymbol{#1\pm#2}$}
\usepackage[
hidelinks,
colorlinks=false,
linkcolor=black, 
citecolor=black,
urlcolor=black 
]{hyperref}

\newcommand{\figref}[1]{\hyperref[#1]{\textcolor{RevisionText}{Fig.~\ref*{#1}}}}
\newcommand{\tabref}[1]{\hyperref[#1]{\textcolor{RevisionText}{Table~\ref*{#1}}}}
\journal{ISPRS Journal of Photogrammetry and Remote Sensing}

\begin{document}

\begin{frontmatter}
\title{The First Assessment of PhiSat-2 Imagery for Monocular Building Height Estimation}

\author[first]{Yanjiao Song}
\ead{yanjiaosong@whu.edu.cn}
\author[first]{Bowen Cai}
\author[first,second]{Timo Balz\corref{cor1}}
\ead{balz@whu.edu.cn}
\author[first]{Zhenfeng Shao}
\author[third]{Neema Simon Sumari}
\author[fourth]{James Magidi}
\author[fifth]{Walter Musakwa}

\affiliation[first]{organization={State Key Laboratory of Information Engineering in Surveying, Mapping and Remote Sensing (LIESMARS), Wuhan University},
            city={Wuhan},
            postcode={430079}, 
            state={Hubei},
            country={China}}
\affiliation[second]{organization={Department of Geomatics Engineering, Institut Teknologi Sepuluh Nopember},
         	city={Surabaya},
         	postcode={60111}, 
         	state={East Java},
         	country={Indonesia}}
\affiliation[third]{organization={Department of Informatics and Information Technology, College of Natural and Applied Sciences (CoNAS), Sokoine University of Agriculture},
	city={Morogoro},
	postcode={3038}, 
	state={Morogoro},
	country={Tanzania}}
\affiliation[fourth]{organization={Geomatics Department, Tshwane University of Technology},
	city={Pretoria},
	postcode={0183}, 
	state={Gauteng},
	country={South Africa}}
\affiliation[fifth]{organization={Department of Geography, Environmental Management and Energy Studies, Faculty of Science, University of Johannesburg},
	city={Johannesburg},
	postcode={2006}, 
	state={Gauteng},
	country={South Africa}} 	
	
\cortext[cor1]{Corresponding author}

\begin{abstract}

\rev{PhiSat-2 provides multispectral observations at a 4.75\,m spatial resolution across geographically diverse scenes, offering a new source of remote sensing imagery for monocular building-height estimation.} \rev{To comprehensively assess its utility, this study introduces the PhiSat-2--Height Dataset (PHDataset) and the Two-Stream Ordinal Network (TSONet).} PHDataset integrates global PhiSat-2 imagery with open building-height references and contains 9,475 co-registered patch pairs from 26 cities worldwide. \rev{In TSONet, multiscale decoupled height representation (MDHR) separates adaptive bin generation from spatial assignment, while footprint-conditioned height reconstruction (FCHR) incorporates footprint guidance at feature and reconstruction levels.} \rev{Under city-disjoint evaluation, TSONet reduces building RMSE and MAE by 3.64\% and 4.49\%, respectively, relative to the most accurate comparison method originally designed for building-height estimation, while maintaining similar height-derived footprint IoU.} \rev{Ablations support the combined design for height accuracy and footprint consistency.} \rev{PhiSat-2 reduces building RMSE by 3.73\% and 6.05\% relative to the evaluated Sentinel-2 and Landsat 8 configurations, respectively, with all seven bands providing the lowest building-height errors among its tested band subsets.} \rev{LiDAR-based testing supports PHDataset-style supervision in a held-out city, while selected patch comparisons show finer urban height patterns than coarser global products.} \rev{These findings highlight the value of PhiSat-2 spatial–spectral information for urban building height mapping, with PHDataset and TSONet providing a foundation for further development.} \rev{PHDataset is available at \url{https://huggingface.co/datasets/Yanjiao-WHU/PHDataset}, and the code at \url{https://github.com/SpectorSong/TSONet}.}

\end{abstract}


\end{frontmatter}

\section{Introduction}
Accurate building height estimation (BHE) is fundamental to characterizing the built environment, as it provides critical vertical information for a more complete representation of urban morphology, including attributes such as building volume and floor area ratio~\citep{li2020developing, cai2023deep}. Reliable building height data support a wide range of applications, including energy-use analysis~\citep{godoy2018energy}, population estimation~\citep{tomas2016urban}, climate adaptation~\citep{li2025building}, and disaster risk assessment~\citep{tian2025fire}. However, fine-grained and up-to-date building height information is still unevenly available across regions, limiting its use for large-area urban analysis. This has driven increasing interest in estimating building height from satellite remote sensing imagery, which provides an efficient and scalable source of observations for large-area urban mapping~\citep{hong2026dual}. \par
A variety of remote sensing approaches have been explored for BHE. Stereo optical imagery has been widely used to reconstruct three-dimensional urban information~\citep{chen2023leveraging}, but it is often difficult to acquire such data with both broad spatial coverage and high temporal frequency. Spaceborne LiDAR observations provide highly accurate vertical measurements~\citep{ma2023mapping}, yet their sparse sampling pattern limits their suitability for high-resolution wall-to-wall mapping. Synthetic aperture radar (SAR) has also been extensively used for BHE because of its sensitivity to geometric structure, including single-SAR methods~\citep{recla2024sar2height}, interferometric SAR (InSAR) methods~\citep{zhu2023algorithm, afzal2025}, and tomographic SAR (TomoSAR) methods~\citep{chen2026reconstructing}. Nevertheless, SAR imagery is often affected by speckle noise, side-looking distortions such as layover and shadows, and limited spectral information, which restrict its ability to comprehensively characterize complex buildings~\citep{frantz2021national, liu2024softformer}. To address these limitations, some studies have explored SAR--optical fusion for BHE~\citep{wu2023first, cao2024deep}. However, the effectiveness of such fusion strategies is still constrained by temporal inconsistency and geometric misalignment between multimodal data sources, limiting their applicability in large-scale operational mapping.\par
Motivated by advances in monocular depth estimation (MDE), recent studies have increasingly explored monocular height estimation (MHE) from optical imagery~\citep{esfahani2025height}. Compared with stereo-based reconstruction and multimodal fusion approaches, MHE requires only a single optical image as input, making it a flexible and scalable solution for large-area BHE. Nevertheless, MHE remains inherently ill-posed because building height cannot be directly measured from monocular optical imagery, but must instead be inferred from indirect visual cues~\citep{zhao2023semantic, chen2024heightformer}. This task is further complicated by the ambiguous relationship between optical appearance and building height, the heterogeneity of building forms and surrounding urban contexts, and the long-tailed distribution of building heights~\citep{chen2023adaptive, zhang2025morphological}. Useful cues may include roof size and shape, shadow patterns, texture, spectral contrast, and local urban morphology, but these cues can be weakened by illumination differences, occlusion, background clutter, and spectral confusion with other land-cover classes~\citep{srivastava2017joint, song2026enhancing}. Therefore, the key issue is not whether monocular optical imagery can deterministically recover building height, but whether a specific sensor provides sufficient spatial--spectral information to support useful empirical height estimation under clearly defined data and label conditions.\par

\rev{PhiSat-2 (also written as $\Phi$Sat-2) provides seven visible-to-near-infrared multispectral bands at a nominal ground sampling distance of 4.75\,m.} \rev{Its combination of spatial detail and spectral coverage motivates an assessment of building-height estimation from a single image.} \rev{Three questions motivate this assessment: how well PhiSat-2 supports BHE relative to other optical inputs, how a task-specific network can jointly model height and building-region structure, and how performance varies across unseen cities and reference sources.} \rev{These questions call for a geographically separated dataset together with complementary sensor and LiDAR comparisons.}

\rev{To address these questions, the \textit{PhiSat-2--Height Dataset (PHDataset)} is constructed and the \textit{Two-Stream Ordinal Network (TSONet)} is proposed.} \rev{PHDataset links observations from 26 cities to open building-height references and provides a fixed city-disjoint evaluation.} \rev{TSONet combines multiscale decoupled height representation (MDHR) with footprint-conditioned height reconstruction (FCHR).} \rev{MDHR separates image-adaptive bin generation from spatial assignment, while FCHR conditions height features locally and uses patch-level footprint information for continuous height reconstruction.} \rev{The term ``ordinal'' denotes the ordered bins; learning is driven by continuous height regression and auxiliary footprint supervision.} \rev{The dense formulation estimates height and its spatial support on a common grid without requiring individual-building delineation at inference.} \rev{The contributions are threefold:}
\begin{enumerate}[label=(\arabic*), leftmargin=2.2em, itemindent=0pt, align=left]
\item \rev{PHDataset is introduced, comprising 9,475 image--height-reference patch pairs from 30 PhiSat-2 scenes across 26 cities.} \rev{Its city-disjoint partition and explicit reference provenance provide a reproducible basis for evaluating building-height estimation from this sensor.}
\item \rev{TSONet is proposed to integrate decoupled bin generation and spatial assignment with footprint guidance at both feature and reconstruction levels.} \rev{The roles of MDHR and FCHR within the complete design are assessed through ablation experiments using both height accuracy and height-derived spatial scores.}
\item \rev{A first task-specific assessment of PhiSat-2 for monocular BHE is presented through model comparisons, sensor and input analyses, and evaluations across cities and reference sources, including complementary LiDAR assessments.}
\end{enumerate}

\begin{table}[t]
	\centering
	\caption{Band information of PhiSat-2 used in this study.}
	\label{tab:phisat_2}
	\setlength{\tabcolsep}{6pt}
	\setlength{\heavyrulewidth}{1.2pt}
	\renewcommand{\arraystretch}{1.15}
	\begin{tabular}{c>{\centering\arraybackslash}p{3.0cm}cc}
		\toprule
		Band & Spectral region & Central wavelength (nm) & Bandwidth (nm)  \\
		\specialrule{1.2pt}{0pt}{0pt}
		\midrule
		MS1 & Blue & 490 & 65  \\
		MS2 & Green & 560 & 35  \\
		MS3 & Red & 665 & 30  \\
		MS4 & Red edge & 705 & 15  \\
		MS5 & Red edge & 740 & 15  \\
		MS6 & Red edge--NIR & 783 & 20  \\
		MS7 & NIR & 842 & 115  \\
		\bottomrule
	\end{tabular}
\end{table}

\section{Related work}
\subsection{Monocular height estimation}
\rev{Existing MHE methods include direct regression and bin-based formulations, with the latter encompassing ordinal regression and adaptive-bin classification--regression approaches.} \rev{Direct regression methods typically estimate continuous height values using encoder--decoder architectures.} Mou and Zhu~\citep{mou2018im2height} were among the first to model the relationship between a single optical image and a digital surface model (DSM) using a fully convolutional--deconvolutional network, without introducing any auxiliary supervision. Amirkolaee and Arefi~\citep{amirkolaee2019height} integrated U-Net and ResNet to build a deeper architecture and achieved promising results on three public datasets. Xing et al.~\citep{xing2021gated} further enhanced encoder--decoder interaction through a gated feature aggregation module and introduced multi-scale supervision for decoder features. \par
\rev{Bin-based methods discretize the height range into intervals.} \rev{Ordinal regression explicitly models ordered labels, whereas adaptive-bin classification--regression methods reconstruct continuous values from bin centers and their predicted probabilities.} \rev{This strategy has been shown to improve optimization stability and estimation quality in MDE, and has gradually been introduced into MHE.} Similar to DORN~\citep{fu2018deep} in MDE, Li et al.~\citep{li2020height} were the first to introduce an ordinal encoder--decoder framework into MHE, transforming feature embeddings into discrete height maps during decoding and optimizing the network with an ordinal loss. Feng et al.~\citep{feng2022soft} further proposed a soft-weighted loss to better convert discrete height intervals into continuous height values by incorporating soft probability distributions. Inspired by AdaBins~\citep{bhat2021adabins}, which adaptively generates depth bins in MDE, HTC-DC Net~\citep{chen2023htc} formulates MHE as a classification--regression task based on a vision transformer (ViT)~\citep{dosovitskiy2020image}, while introducing several optimizations tailored to the distribution of height samples. \par
\rev{Bin-based estimation requires coordination between image-level bin generation and pixel-wise representations.} \rev{BinsFormer~\citep{li2024binsformer} refines bin queries across scales, generating adaptive bin widths and obtaining spatial-assignment scores through dot products between bin embeddings and pixel features.} \rev{In TSONet, queries determine bin locations, whereas an independent convolutional head produces pixel-wise bin probabilities from footprint-conditioned height features.}\par

\subsection{Segmentation-guided height estimation}
Semantic segmentation and height estimation are closely related dense prediction tasks, because semantic categories, object extents, and building boundaries provide important contextual constraints for height inference. \rev{Early multi-task studies introduced semantic segmentation as an auxiliary task for monocular height estimation, showing that semantic context can help height estimation by improving object-level discrimination and boundary preservation~\citep{srivastava2017joint}.} Following this idea, several methods have explored more explicit task interaction mechanisms. Liu et al.~\citep{liu2022associatively} proposed ASSEH, which uses a shared backbone, task-specific gating units, cross-task propagation, and task-specific decoders to jointly perform semantic segmentation and height estimation. Gao et al.~\citep{gao2023joint} further introduced cross-task and cross-pixel contrastive losses to enhance consistency between segmentation and height representations. Zhao et al.~\citep{zhao2023semantic} incorporated semantic supervision into unsupervised domain adaptation for single-view height estimation, using semantic consistency and cross-task attention to improve height feature learning. \rev{In a related multi-task setting, HTC-DC Net~\citep{chen2023htc} supervises foreground classification using a 1\,m height threshold and selects foreground or background bin probabilities for height reconstruction according to whether the foreground probability exceeds 0.5.} More recently, FusedSeg-HE~\citep{gultekin2025fusing} added an auxiliary segmentation head to a CNN--ViT fused height estimation framework, improving pixel-level alignment especially around object boundaries. These studies indicate that segmentation- or semantics-related supervision can provide useful structural guidance for height estimation by constraining object extent, suppressing background interference, and improving boundary consistency.\par
However, most existing segmentation-guided MHE methods have been developed and evaluated on high-resolution aerial imagery or established benchmark datasets, and their suitability for newly available PhiSat-2 imagery remains unclear. This issue is particularly relevant for PhiSat-2 because its 4.75\,m resolution provides more spatial detail than coarser open optical sensors, but is still limited for precise pixel-level building boundary delineation. \rev{For PhiSat-2, footprint guidance is incorporated into both local height features and continuous height reconstruction through feature conditioning and a patch-shared weight derived from footprint probabilities.\par}

\section{Materials}
To evaluate the capability of PhiSat-2 imagery for BHE, global PhiSat-2 satellite images and open-source building height labels were combined to construct a new dataset, termed \textit{PHDataset}. Data from several sources were first collected and filtered, and then integrated and processed into a machine learning dataset. \par

\subsection{PhiSat-2 imagery}
\label{sec:phisat_2}
According to the ESA mission overview technical note \citep{Longepe2023PhiSat2}, PhiSat-2 is equipped with the MultiScape100 visible-to-near-infrared (VNIR) imager, which provides eight spectral bands in total, including seven multispectral bands and one additional panchromatic band, with a ground sampling distance of 4.75\,m and a swath width of 19.4\,km at the reference altitude of 500\,km. In this study, only the seven multispectral bands were used, as listed in \tabref{tab:phisat_2}. Since the payload adopts a push-broom imaging mode, in which different spectral bands are acquired separately along track, practical PhiSat-2 products may exhibit residual geometric inconsistency, including spatial offsets and inter-band misregistration; these issues will be further discussed in Section~\ref{sec:geometric_sensitivity}.\par

\begin{figure*}[t]
	\centering 
	\includegraphics[width=1.0\textwidth, angle=0]{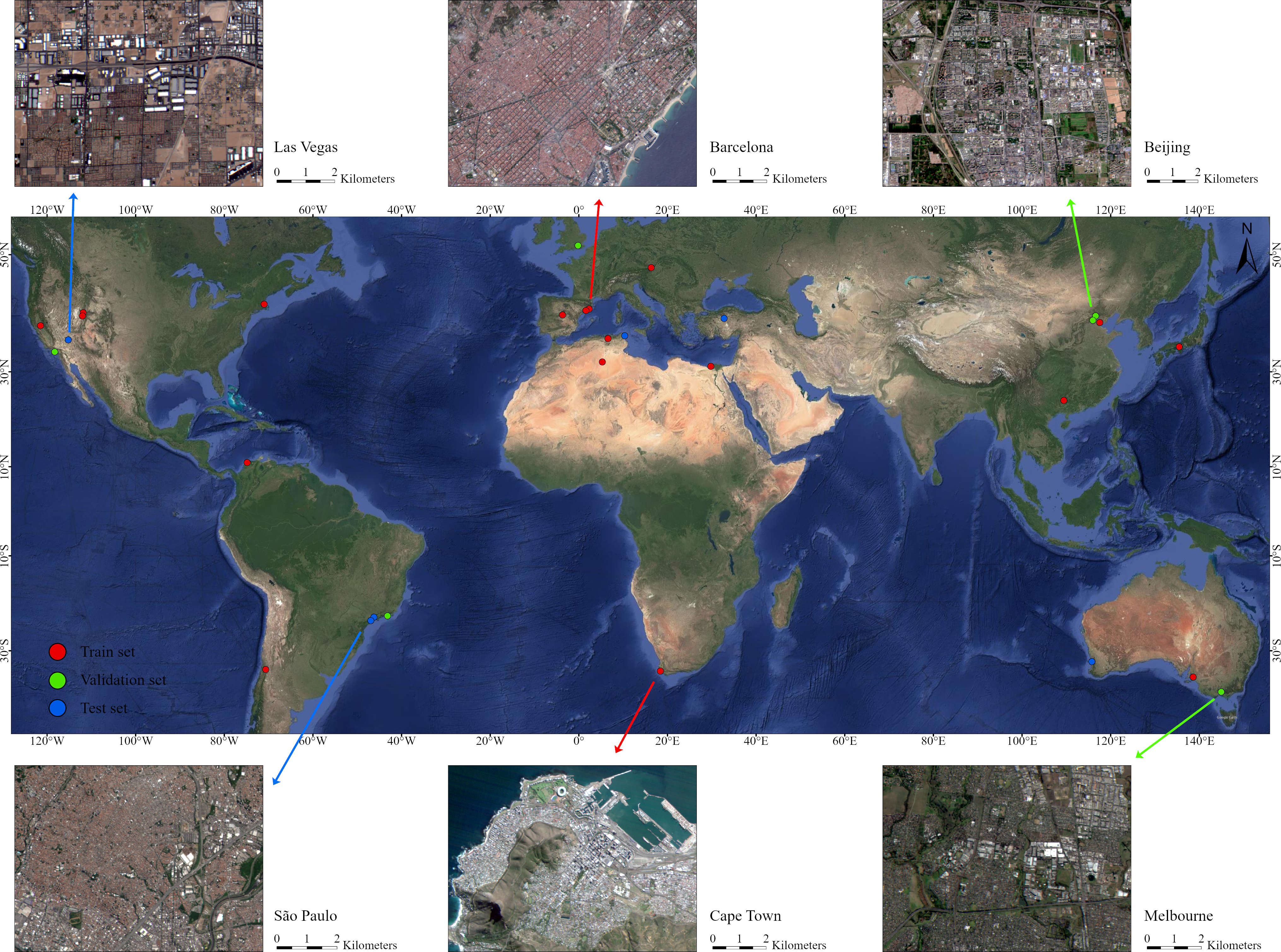}	
	\caption{\rev{Spatial distribution of the selected PhiSat-2 scenes and representative imagery examples.}} 
	\label{fig:samples}
\end{figure*}

\begin{table}[tp]
\centering
\caption{\rev{Scene characteristics and city-disjoint partition of PHDataset.}}
\label{tab:dataset_inventory}
\footnotesize
\setlength{\tabcolsep}{3pt}
\renewcommand{\arraystretch}{1.12}
\begin{tabular*}{\textwidth}{@{\extracolsep{\fill}}cccccccc@{}}
\toprule
\rev{City} & \rev{\shortstack{Acquisition\\date (UTC)}} & \rev{Source} & \rev{\shortstack{Reference\\year/period}} & \rev{Split} & \rev{Patches} & \rev{\shortstack{$B$ (\%)}} & \rev{\shortstack{$\mu_h\pm\sigma_h$ (m)}}  \\
\midrule
\rev{Adelaide} & \rev{2025-08-03} & \rev{M} & \rev{2014--2024} & \rev{Train} & \rev{305} & \rev{10.99} & \rev{$5.36\pm2.09$}  \\
\rev{Alexandria} & \rev{2025-08-31} & \rev{G} & \rev{2020} & \rev{Train} & \rev{237} & \rev{9.37} & \rev{$11.86\pm6.29$}  \\
\rev{Ankara} & \rev{2025-08-23} & \rev{G} & \rev{2020} & \rev{Test} & \rev{394} & \rev{6.63} & \rev{$18.62\pm11.76$}  \\
\rev{Barcelona} & \rev{2025-05-25} & \rev{M} & \rev{2014--2024} & \rev{Train} & \rev{192} & \rev{24.64} & \rev{$11.98\pm4.86$}  \\
\rev{Barcelona} & \rev{2025-06-24} & \rev{M} & \rev{2014--2024} & \rev{Train} & \rev{441} & \rev{5.68} & \rev{$7.71\pm3.70$}  \\
\rev{Barranquilla} & \rev{2025-04-25} & \rev{G} & \rev{2020} & \rev{Train} & \rev{124} & \rev{21.07} & \rev{$11.12\pm6.81$}  \\
\rev{Beijing} & \rev{2025-04-28} & \rev{G} & \rev{2020} & \rev{Val.} & \rev{485} & \rev{12.56} & \rev{$16.18\pm10.66$}  \\
\rev{Beijing} & \rev{2025-06-15} & \rev{G} & \rev{2020} & \rev{Val.} & \rev{389} & \rev{12.39} & \rev{$12.79\pm7.94$}  \\
\rev{Boston} & \rev{2025-08-13} & \rev{M} & \rev{2014--2024} & \rev{Train} & \rev{262} & \rev{11.53} & \rev{$9.58\pm5.37$}  \\
\rev{Cape Town} & \rev{2025-05-05} & \rev{G} & \rev{2020} & \rev{Train} & \rev{341} & \rev{6.34} & \rev{$8.42\pm6.34$}  \\
\rev{Constantine} & \rev{2025-10-30} & \rev{G} & \rev{2020} & \rev{Train} & \rev{291} & \rev{5.36} & \rev{$10.99\pm6.27$}  \\
\rev{Las Vegas} & \rev{2025-08-14} & \rev{M} & \rev{2014--2024} & \rev{Test} & \rev{354} & \rev{11.47} & \rev{$5.40\pm2.42$}  \\
\rev{Liuzhou} & \rev{2025-10-28} & \rev{G} & \rev{2020} & \rev{Train} & \rev{235} & \rev{6.90} & \rev{$23.66\pm8.77$}  \\
\rev{London} & \rev{2025-04-06} & \rev{M} & \rev{2014--2024} & \rev{Val.} & \rev{500} & \rev{19.71} & \rev{$8.46\pm3.97$}  \\
\rev{Los Angeles} & \rev{2025-09-03} & \rev{M} & \rev{2014--2024} & \rev{Val.} & \rev{272} & \rev{18.63} & \rev{$5.32\pm2.32$}  \\
\rev{Madrid} & \rev{2024-12-22} & \rev{M} & \rev{2014--2024} & \rev{Train} & \rev{495} & \rev{16.69} & \rev{$10.96\pm4.87$}  \\
\rev{Melbourne} & \rev{2025-06-23} & \rev{M} & \rev{2014--2024} & \rev{Val.} & \rev{217} & \rev{16.07} & \rev{$5.37\pm1.82$}  \\
\rev{Osaka} & \rev{2025-04-14} & \rev{G} & \rev{2020} & \rev{Train} & \rev{500} & \rev{20.10} & \rev{$16.68\pm14.73$}  \\
\rev{Ouargla} & \rev{2025-10-26} & \rev{G} & \rev{2020} & \rev{Train} & \rev{252} & \rev{4.87} & \rev{$8.18\pm3.65$}  \\
\rev{Perth} & \rev{2025-06-26} & \rev{M} & \rev{2014--2024} & \rev{Test} & \rev{228} & \rev{10.81} & \rev{$5.10\pm1.47$}  \\
\rev{Rio de Janeiro} & \rev{2025-05-02} & \rev{G} & \rev{2020} & \rev{Val.} & \rev{193} & \rev{12.11} & \rev{$22.96\pm16.10$}  \\
\rev{Sacramento} & \rev{2025-08-25} & \rev{M} & \rev{2014--2024} & \rev{Train} & \rev{382} & \rev{16.38} & \rev{$5.25\pm2.16$}  \\
\rev{Salt Lake City} & \rev{2025-08-06} & \rev{M} & \rev{2014--2024} & \rev{Train} & \rev{425} & \rev{8.24} & \rev{$5.40\pm1.83$}  \\
\rev{Salt Lake City} & \rev{2025-08-10} & \rev{M} & \rev{2014--2024} & \rev{Train} & \rev{210} & \rev{1.37} & \rev{$5.31\pm1.69$}  \\
\rev{Santiago} & \rev{2025-06-18} & \rev{G} & \rev{2020} & \rev{Train} & \rev{371} & \rev{15.15} & \rev{$7.73\pm4.59$}  \\
\rev{S\~ao Paulo} & \rev{2025-03-22} & \rev{G} & \rev{2020} & \rev{Test} & \rev{265} & \rev{35.66} & \rev{$11.62\pm7.92$}  \\
\rev{S\~ao Paulo} & \rev{2025-03-27} & \rev{G} & \rev{2020} & \rev{Test} & \rev{241} & \rev{32.11} & \rev{$12.00\pm8.25$}  \\
\rev{Tianjin} & \rev{2025-04-06} & \rev{G} & \rev{2020} & \rev{Train} & 490 & \rev{9.20} & \rev{$11.53\pm7.17$}  \\
\rev{Tunis} & \rev{2025-08-15} & \rev{G} & \rev{2020} & \rev{Test} & \rev{238} & \rev{10.10} & \rev{$9.68\pm4.92$}  \\
\rev{Vienna} & \rev{2025-04-27} & \rev{E} & \rev{2013} & \rev{Train} & \rev{146} & \rev{15.56} & \rev{$14.16\pm9.18$}  \\
\bottomrule
\end{tabular*}
\par\medskip
\begin{minipage}{\textwidth}
\footnotesize
\rev{M: Microsoft Global ML Building Footprints and associated heights; E: EUBUCCO v0.1; G: 3D-GloBFP.} \rev{$B$ denotes the proportion of jointly valid image--reference pixels with reference heights above 2\,m.} \rev{$\mu_h$ and $\sigma_h$ are the corresponding pixel-pooled mean and population standard deviation across retained patches, including overlapping locations.}\end{minipage}
\end{table}

\rev{PhiSat-2 imagery was collected for urban areas worldwide; the retained acquisitions span 22 December 2024 to 30 October 2025, as documented in \tabref{tab:dataset_inventory}.} \rev{Thirty scenes from 26 cities were retained, as shown in \figref{fig:samples}, after screening for cloud cover and geometric quality.} The seven multispectral bands were composited and reprojected to the WGS 1984 Web Mercator projection (EPSG:3857). \rev{The selected images were manually georeferenced against high-resolution Google Earth imagery using corresponding landmarks, with a common scene-specific transformation applied to all seven bands.} \rev{The processed imagery was resampled onto an EPSG:3857 grid with a pixel spacing of 4.75\,m.} \rev{Residual image-to-reference alignment was assessed after preprocessing using 15 manually identified landmark pairs per scene, giving 450 pairs across all 30 scenes.} \rev{Landmark selection was guided by 15 spatial strata within each scene.} \rev{For $N_p$ landmark pairs, the pairwise displacement $d_i$ and the two-dimensional root mean squared error (RMSE) are defined as}
{\color{RevisionText}
\begin{equation}
\label{eq:registration_rmse}
\begin{aligned}
d_i&=\sqrt{(\Delta x_i)^2+(\Delta y_i)^2}\\
\mathrm{RMSE}_{\mathrm{reg}}&=\sqrt{\frac{1}{N_p}\sum_{i=1}^{N_p}d_i^2}
\end{aligned}
\end{equation}
}%
\noindent \rev{Here, $\Delta x_i$ and $\Delta y_i$ are the column and row coordinate differences of the $i$th corresponding landmark pair, expressed in pixels on the final image grid, and $N_p$ is the total number of pairs in the assessment.} \rev{The median displacement is the median of the individual $d_i$ values.} \rev{Across all 450 pairs, the pooled RMSE was 3.81 pixels and the median displacement was 2.26 pixels.} \rev{These statistics describe residual agreement with the Google Earth reference and may include landmark-localization uncertainty associated with differences in spatial resolution and imaging geometry.\par}
\rev{To support the cross-sensor comparison, corresponding Sentinel-2 Level-1C and Landsat 8 Collection 2 Tier 1 top-of-atmosphere images were obtained through Google Earth Engine for the study areas.} \rev{The Sentinel-2 inputs use visible bands B2--B4 and near-infrared B8, with red-edge bands B5--B7 included in the seven-band configuration.} \rev{The Landsat 8 inputs use visible bands B2--B4 and near-infrared B5.} \rev{The prepared Sentinel-2 and Landsat 8 source grids have 10\,m and 30\,m spacing, respectively; Sentinel-2 red-edge bands have a native resolution of 20\,m.} \rev{These data are geospatially resampled to the PhiSat-2 patch grid for the cross-sensor comparison, without panchromatic fusion.} \rev{The acquired sensor configurations retain differences in spectral response and acquisition timing.\par}

\subsection{Building height references}
\label{sec:height_reference}
Dense airborne LiDAR and high-resolution DSM/DTM products can provide accurate physical height references~\citep{chen2023htc,recla2024sar2height}, but their spatial availability is highly uneven across countries and cities. Map-derived building attributes, such as A-map floor-number records for major Chinese cities, can also be converted into approximate building heights~\citep{LI2020111859,cai2023deep}, but this conversion is affected by variable floor heights and roof structures, and temporal inconsistency with PhiSat-2 observations. Moreover, many open LiDAR, DSM, and map-derived products were produced many years before the PhiSat-2 observations. \rev{Such temporal inconsistency may introduce apparent height errors caused by urban redevelopment, rather than by errors in model-based height estimation.}\par

\rev{Three open products provide building-height references according to regional availability.} \rev{Microsoft Global ML Building Footprints and associated height estimates\footnote{\url{https://github.com/microsoft/GlobalMLBuildingFootprints}} cover North America, Australia, and most European study cities.} \rev{The product is derived from multi-temporal Bing Maps imagery acquired between 2014 and 2024.} \rev{Vienna uses EUBUCCO v0.1~\citep{milojevic2023eubucco}, whose local source is the municipal roof model based on a 2013 surface model\footnote{\url{https://www.wien.gv.at/stadtplanung/generalisiertes-dachmodell-daten}}.} \rev{Cities in Asia, South America, and Africa use 3D-GloBFP~\citep{che20243d}, a building-scale height product with a nominal reference year of 2020.} \rev{These source epochs describe heterogeneous reference products and do not imply contemporaneous building surveys.\par}

\rev{All height references were projected to EPSG:3857, rasterized at the pixel spacing of the processed PhiSat-2 imagery, and manually aligned to the images using corresponding building landmarks.} \rev{Reference heights are expressed in meters.} \rev{An additional alignment assessment used 15 image--label landmark pairs in each of Cape Town, Osaka, Santiago, London, and Vienna, representing the three reference sources.} \rev{The 75 image--label pairs yielded a pooled RMSE of 2.32 pixels and a median displacement of 1.45 pixels.} \rev{For separately selected image--Google Earth landmarks in the same scenes, the corresponding values were 6.26 and 3.09 pixels.} \rev{These comparisons characterize residual spatial correspondence in the assessed scenes; differences in image resolution and landmark localization also contribute to the observed displacements.\par}
\rev{The source footprints define the spatial support of the rasterized height references.} \rev{Deriving the auxiliary building mask from the processed height raster ensures that footprint and height supervision share the same pixel grid, validity support, and building-region definition, without introducing a separate rasterization of the source polygons.} \rev{The mask is defined as}
{\color{RevisionText}
\begin{equation}
\label{eq:reference_footprint}
f_i=\mathbf{1}(h_i>\tau)
\end{equation}
}%
\noindent \rev{Here, $h_i$ is the reference height at pixel $i$, $f_i$ is the corresponding auxiliary footprint target, $\mathbf{1}(\cdot)$ is the indicator function, and $\tau=2\,$m.} \rev{The 2\,m cutoff excludes near-ground height fluctuations while remaining below the approximately 3\,m single-storey height assumed in previous building-height studies~\citep{cai2023deep,wang2024mf}; it is an operational label definition rather than a universal minimum building height.} \rev{Thresholds of 1 and 3\,m bracket this choice and test its influence on label coverage.} \rev{With the reference heights and valid-data support held fixed across all 9,475 patches, the foreground pixel count increases by 0.45\% at 1\,m and decreases by 1.03\% at 3\,m relative to 2\,m.} \rev{This sensitivity analysis concerns the reference labels and does not measure height-estimation accuracy after retraining.\par}
\rev{To complement the open-reference evaluation, LiDAR-derived normalized digital surface models (nDSMs) provide separate height references in London, Barcelona, and Madrid.} \rev{These three cities do not cover regions labeled with 3D-GloBFP.} \rev{For London, an nDSM was derived from the UK Environment Agency 1\,m LiDAR composite DSM/DTM\footnote{\url{https://developers.google.com/earth-engine/datasets/catalog/UK_EA_ENGLAND_1M_TERRAIN_2022}}, assembled from surveys conducted during 1999--2018.} \rev{For Barcelona and Madrid, the 2016 PNOA-LiDAR building nDSM\footnote{\url{https://centrodedescargas.cnig.es/CentroDescargas/buscar-mapa}} at 2.5\,m resolution was used.} \rev{These products were reprojected, co-registered, and sampled on the PhiSat-2 grid.} \rev{They were not used to construct the main PHDataset labels; they provide the LiDAR evaluation reference and the alternative targets for the dedicated supervision experiment in Section~\ref{sec:lidar_assessment}.\par}

\subsection{Dataset construction}
\label{sec:dataset_construction}

\rev{Temporal differences and residual geometric errors can produce mismatches between the imagery and reference labels.} \rev{Scene-level quality masks delimit areas retained for patch extraction; within those areas, pixel-level validity requires a valid height reference and valid observations in all seven optical bands.} \rev{Valid background pixels remain in the training and evaluation support.\par}
\rev{Co-registered image and reference data were divided into $256\times256$-pixel patches at a stride of 256 pixels.} \rev{Patches at the outer scene boundaries were aligned to the image extent, introducing local overlap.} \rev{Fully invalid patches were excluded, partially valid patches were subsampled, and up to 500 patches were retained per scene.} \rev{PHDataset contains 9,475 patch pairs from 30 scenes across 26 cities.} \par
\rev{The PHDataset uses a city-disjoint partition that assigns all observations from each city exclusively to training, validation, or testing.} \rev{The subsets contain 16, 5, and 5 cities, comprising 5,699, 2,056, and 1,720 patches, respectively.} \rev{The allocation considers geographic coverage, reference sources, building coverage, and height distributions.} \rev{The validation cities are Beijing, London, Los Angeles, Melbourne, and Rio de Janeiro; the test cities are Ankara, Las Vegas, Perth, S\~ao Paulo, and Tunis.} \rev{City assignments are fixed, so patches from the same city or scene, including locally overlapping patches, cannot occur in different subsets.} \rev{\tabref{tab:dataset_inventory} summarizes scene acquisition dates, reference sources and epochs, subset assignments, and patch-level sample composition.} \rev{For PhiSat-2 inputs, min--max normalization parameters are estimated from valid training pixels and applied unchanged to the training, validation, and test subsets.} \par
\rev{The frozen-model LiDAR comparison uses Barcelona and Madrid from the main training subset and London from validation.} \rev{A separate supervision-source experiment defines a spatially separated Barcelona/Madrid training--validation subset and reserves London for testing, as detailed in Section~\ref{sec:lidar_assessment}.} \rev{The cross-sensor comparison uses Sentinel-2 and Landsat 8 observations at the main patch locations, as described in Section~\ref{sec:sensor_info}.\par}

\section{Method}
\label{sec:method}

PHDataset consists of PhiSat-2 samples collected from diverse urban environments, covering buildings with heterogeneous spatial layouts, spectral responses, and height distributions. These characteristics make monocular BHE challenging because height cues in optical imagery are indirect and closely entangled with building footprints, roof appearance, shadows, and surrounding urban context. \rev{To better exploit the structural relationship between footprint delineation and height estimation while representing continuous heights with ordered adaptive bins, a Two-Stream Ordinal Network (\textit{TSONet}) is proposed for BHE.} \rev{TSONet combines multiscale decoupled height representation (MDHR), which separates ordered-bin generation from pixel-wise assignment, with footprint-conditioned height reconstruction (FCHR), which integrates footprint features and probabilities into height estimation.}

\subsection{Overall structure}
\label{sec:overall_structure}

\begin{figure}[t]
\centering
\includegraphics[width=\textwidth]{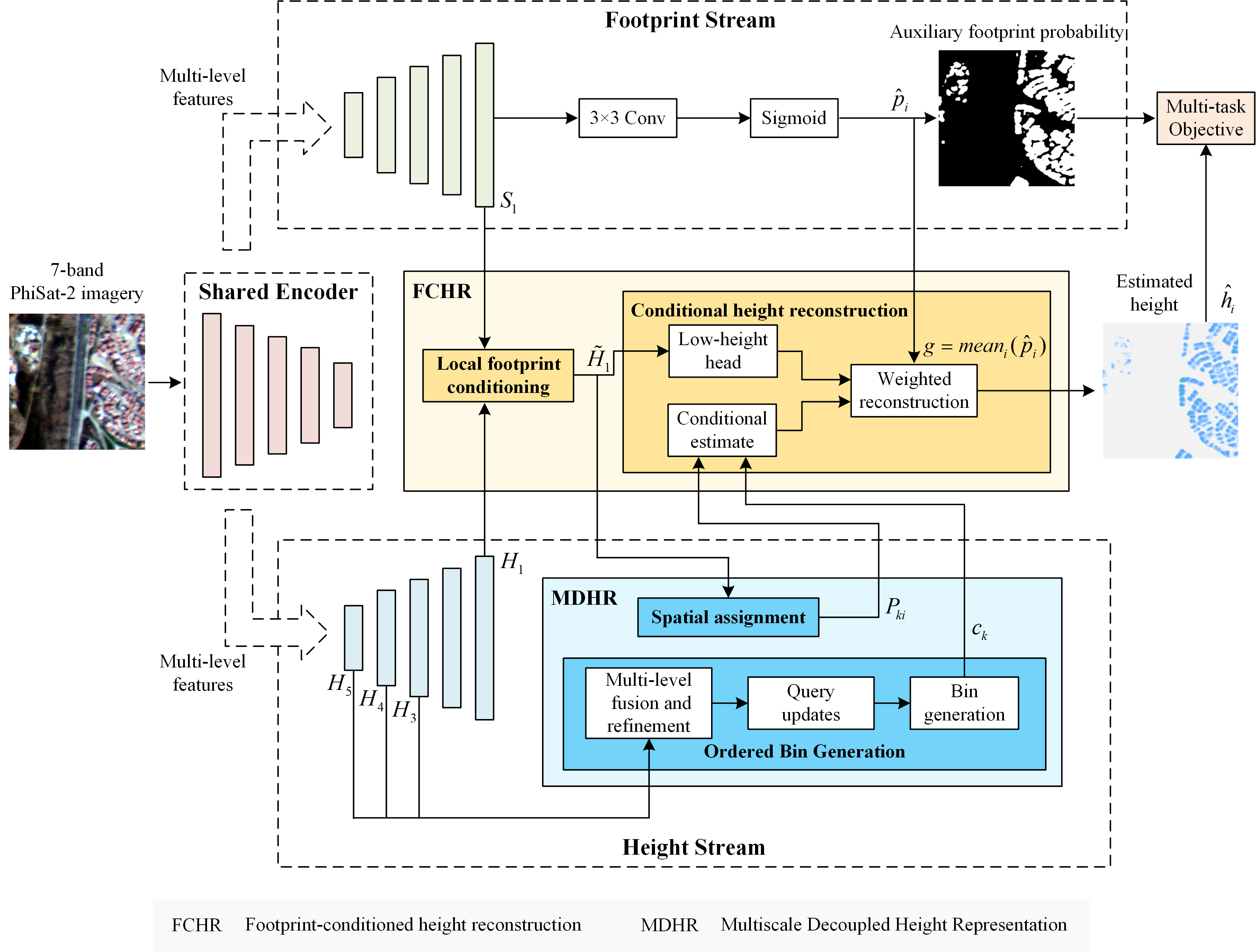}
\caption{\rev{Overall architecture of TSONet.}}
\label{fig:overall_structure}
\end{figure}

\rev{As shown in \figref{fig:overall_structure}, TSONet extracts a shared feature pyramid and decodes it into parallel height and footprint streams.} All seven bands of PhiSat-2 imagery are first fed into the shared encoder, which adopts a five-level hierarchical design analogous to U-Net to extract multi-level feature representations. \rev{For input $\mathbf I\in\mathbb R^{7\times H\times W}$, encoder feature $\mathbf E_\ell$ has $C_\ell=2^{\ell-1}D$ channels and spatial size $(H_\ell,W_\ell)=(H,W)/2^{\ell-1}$, where $\ell=1,\ldots,5$ and $D=64$.} \rev{Let $\operatorname{Conv}_{k\times k,s}^{a\to b}$ denote a convolution with kernel size $k$, stride $s$, and $a$ input and $b$ output channels; stride 1 is omitted below.} \rev{With $\rho$ denoting ReLU and $\operatorname{BN}$ batch normalization, the basic block and its two-layer composition are}
{\color{RevisionText}
\begin{equation}
\label{eq:conv_blocks}
\begin{aligned}
\mathcal B_s^{a\to b}&=\rho\circ\operatorname{BN}\circ\operatorname{Conv}_{3\times3,s}^{a\to b}\\
\mathcal D_s^{a\to b}&=\mathcal B_1^{b\to b}\circ\mathcal B_s^{a\to b}
\end{aligned}
\end{equation}
}%
\noindent \rev{Composition $\circ$ applies operators from right to left, with independent parameters for each layer instance.} \rev{The encoder applies $\mathcal D_1^{7\to D}$ at level 1 and $\mathcal D_2^{C_{\ell-1}\to C_\ell}$ at levels 2--5.} The resulting feature pyramid is subsequently delivered to the two task-specific streams.

\rev{The height and footprint decoders have independent parameters and follow the same coarse-to-fine structure:}
{\color{RevisionText}
\begin{equation}
\label{eq:task_decoders}
\begin{aligned}
\mathbf H_5&=\mathcal P_5^h(\mathbf E_5)\\
\mathbf S_5&=\mathcal P_5^f(\mathbf E_5)\\
\mathbf H_\ell&=\mathcal D_\ell^h\!\left[\operatorname{Cat}\!\left(\operatorname{Up}_\ell(\mathbf H_{\ell+1}),\mathbf E_\ell\right)\right]\\
\mathbf S_\ell&=\mathcal D_\ell^f\!\left[\operatorname{Cat}\!\left(\operatorname{Up}_\ell(\mathbf S_{\ell+1}),\mathbf E_\ell\right)\right]
\end{aligned}
\end{equation}
}%
\noindent \rev{The $1\times1$ projections $\mathcal P_5^h,\mathcal P_5^f$ map $16D$ to $8D$ channels.} \rev{For $\ell=4,3,2,1$, $\operatorname{Up}_\ell$ is bilinear interpolation to the corresponding encoder grid, $\operatorname{Cat}$ is channel-wise concatenation, and $\mathcal D_\ell^h,\mathcal D_\ell^f$ are independent stride-1 instances of Eq.~\eqref{eq:conv_blocks}, producing $8D,4D,2D,D$ channels, respectively.} 

\rev{These decoder outputs are combined by MDHR and FCHR to reconstruct height.} \rev{FCHR first conditions $\mathbf H_1$ with $\mathbf S_1$, producing $\widetilde{\mathbf H}_1$.} \rev{MDHR generates ordered bins from $\mathbf H_5,\mathbf H_4,\mathbf H_3$ and pixel-wise assignment probabilities from this conditioned feature.} \rev{FCHR then combines the resulting bin-based estimate with a complementary estimate from $\widetilde{\mathbf H}_1$, using a patch-shared weight derived from the footprint probabilities.}

\rev{At full resolution, a $3\times3$ convolution maps the footprint feature $\mathbf S_1$ to logits $z_i$.} \rev{Applying sigmoid yields the auxiliary building-membership probabilities $\hat p_i=\sigma(z_i)$.} The proposed framework follows a dense image-to-map formulation rather than an object-level height attribution formulation based on pre-defined building polygons. \rev{During training, auxiliary footprint targets are derived from the height references.} \rev{At inference, both streams remain active and only the input image is required.}

\subsection{\rev{Multiscale decoupled height representation}}
\label{sec:mdhr}

\begin{figure}[t]
\centering
\includegraphics[width=\textwidth]{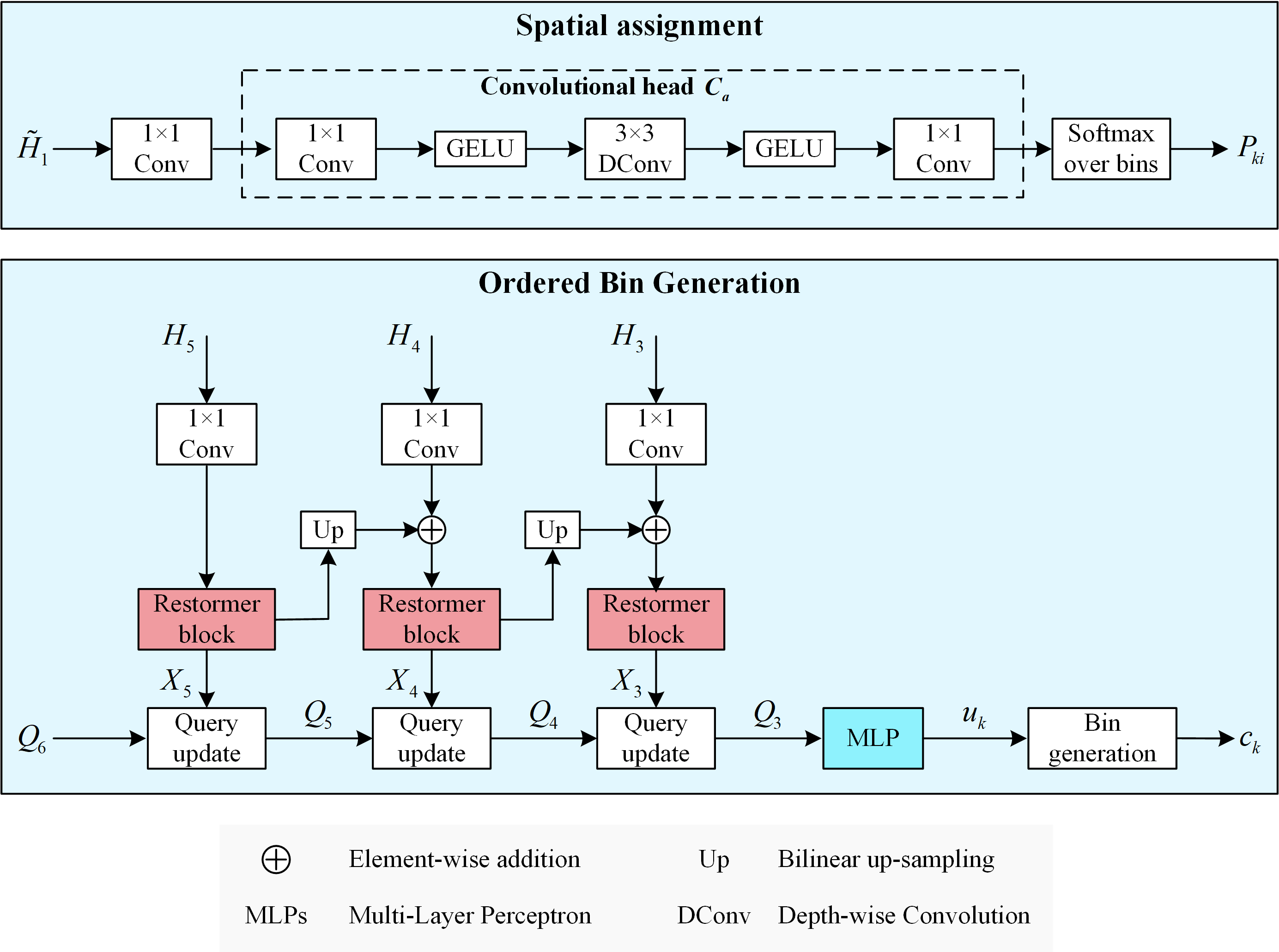}
\caption{\rev{The structure of MDHR (multiscale decoupled height representation).}}
\label{fig:mdhr}
\end{figure}

\rev{MDHR separates image-adaptive ordered-bin generation from pixel-wise spatial assignment, as illustrated in \figref{fig:mdhr}.} \rev{Its bin-generation pathway builds on coarse-to-fine query refinement~\citep{li2024binsformer}, whereas an independent convolutional pathway maps full-resolution conditioned features to assignment probabilities.} \rev{Both pathways are jointly optimized through the final height objective.}

\rev{Height features at levels 5, 4, and 3 are projected to $d=4D$ channels and refined from coarse to fine:}
{\color{RevisionText}
\begin{equation}
\label{eq:multiscale_height}
\begin{aligned}
\mathbf X_5&=\mathcal R_5(\mathcal P_5\mathbf H_5)\\
\mathbf X_\ell&=\mathcal R_\ell\!\left(\mathcal P_\ell\mathbf H_\ell+
\operatorname{Up}_\ell(\mathbf X_{\ell+1})\right),\quad \ell=4,3
\end{aligned}
\end{equation}
}%
\noindent \rev{Here, $\mathcal P_\ell$ is a $1\times1$ projection, and $\operatorname{Up}_\ell$ follows Eq.~\eqref{eq:task_decoders}.} \rev{Each Restormer-style block $\mathcal R_\ell$~\citep{zamir2022restormer} combines pre-normalized eight-head transposed channel attention and a gated depthwise-convolutional feed-forward sublayer, each with a residual connection.} \rev{Query pooling reaches one-quarter input resolution at level 3.}

\rev{The $K=64$ learned queries $\mathbf Q_6\in\mathbb R^{K\times d}$ are updated at levels $\ell=5,4,3$ to obtain $\mathbf Q_5,\mathbf Q_4,\mathbf Q_3$.} \rev{Independent initialization from $\mathcal N(0,0.02^2)$ breaks query symmetry; no positional embeddings are added to queries or feature tokens.} \rev{Flattening $\mathbf X_\ell$ gives $\mathbf Y_\ell\in\mathbb R^{n_\ell\times d}$ with $n_\ell=H_\ell W_\ell$ spatial positions.} \rev{With $\operatorname{clip}_{\kappa}$ denoting element-wise clipping to $[-\kappa,\kappa]$, $\kappa=50$, and query--feature similarity dimension $d_q=32$, each update is}
{\color{RevisionText}
\begin{equation}
\label{eq:query_pooling}
\begin{aligned}
\mathbf A_\ell&=\operatorname{softmax}_{\mathrm{space}}\!\left[
\operatorname{clip}_{\kappa}\!\left(
\frac{(\operatorname{LN}_0(\mathbf Q_{\ell+1})W_Q)
(\operatorname{LN}_0(\mathbf Y_\ell)W_K)^{\mathsf T}}{\sqrt{d_q}}
\right)\right]\\
\mathbf Q_\ell&=\operatorname{LN}\!\left[
\operatorname{LN}_0(\mathbf Q_{\ell+1})+\gamma_\ell
\operatorname{LN}_0(\mathbf A_\ell\mathbf Y_\ell)\right]
\end{aligned}
\end{equation}
}%
\noindent \rev{$\operatorname{softmax}_{\mathrm{space}}$ normalizes each query over spatial positions, so $\mathbf A_\ell\in\mathbb R^{K\times n_\ell}$ has unit row sums.} \rev{$\operatorname{LN}_0$ is channel-wise layer normalization without affine parameters; $\operatorname{LN}$ has learned affine parameters shared across the three updates.} \rev{Shared projections $W_Q,W_K\in\mathbb R^{d\times d_q}$ compute similarity in $d_q$ dimensions, while pooled values retain $d$ channels.} \rev{Each learned scalar $\gamma_\ell=\sigma(a_\ell)$ is initialized to 0.3.}

\rev{A shared MLP maps each row $\mathbf q_{3,k}$ of $\mathbf Q_3$ to a bin-width logit:}
{\color{RevisionText}
\begin{equation}
\label{eq:width_mlp}
\begin{aligned}
\mathcal M&=\operatorname{Linear}_{d\to1}^{(3)}\circ\rho\circ\operatorname{Linear}_{d\to d}^{(2)}\circ\rho\circ\operatorname{Linear}_{d\to d}^{(1)}\\
u_k&=\mathcal M(\mathbf q_{3,k})
\end{aligned}
\end{equation}
}%
\noindent \rev{$\operatorname{Linear}_{a\to b}^{(j)}$ denotes the $j$th affine layer from $a$ to $b$ features.} \rev{For bin generation and assignment, $\mathcal S$ clips logits with $\operatorname{clip}_{\kappa}$, applies softmax over $K$ bins and a probability floor of $10^{-6}$, then renormalizes to unit sum.} \rev{With $h_{\max}=145\,$m, width-ratio floor $\eta=10^{-4}$, and $e_0=0$, the widths, edges, and centers for $k=1,\ldots,K$ are}
{\color{RevisionText}
\begin{equation}
\label{eq:ordered_bins}
\begin{aligned}
r_k&=\frac{\max\{\mathcal S(\mathbf u)_k,\eta\}}
{\sum_{j=1}^{K}\max\{\mathcal S(\mathbf u)_j,\eta\}}\\
w_k&=h_{\max}r_k\\
e_k&=\sum_{j=1}^{k}w_j\\
c_k&=\tfrac12(e_{k-1}+e_k)
\end{aligned}
\end{equation}
}%
\noindent \rev{Positive widths yield increasing edges and centers with $e_K=h_{\max}$; bin locations adapt to each image and are shared across its pixels.} \rev{This ordered representation defines successive height intervals without query sorting or a separate ordinal-classification loss.}

\rev{The complementary spatial-assignment pathway takes the full-resolution feature $\widetilde{\mathbf H}_1$ conditioned by FCHR (Eq.~\eqref{eq:cross_task_context}) and projects it from $D$ to $d$ channels through $\mathcal P_a$.} \rev{The head $\mathcal C_a$ then applies $1\times1$, depthwise $3\times3$, and $1\times1$ convolutions with channel mappings $d\to d\to d\to K$ and GELU after the first two, yielding}
{\color{RevisionText}
\begin{equation}
\label{eq:conv_assignment}
\begin{aligned}
\mathbf L&=\mathcal C_a(\mathcal P_a\widetilde{\mathbf H}_1)\\
P_{ki}&=\mathcal S(\mathbf L_{:i})_k
\end{aligned}
\end{equation}
}%
\noindent \rev{These convolutions use stride 1, and normalization over bins gives $\sum_kP_{ki}=1$ at each pixel.} \rev{Spatial assignment uses neither query--pixel multiplication nor a learned logit temperature; queries determine only bin locations through Eq.~\eqref{eq:ordered_bins}.}

\subsection{\rev{Footprint-conditioned height reconstruction}}
\label{sec:fchr}

\begin{figure}[t]
\centering
\includegraphics[width=\textwidth]{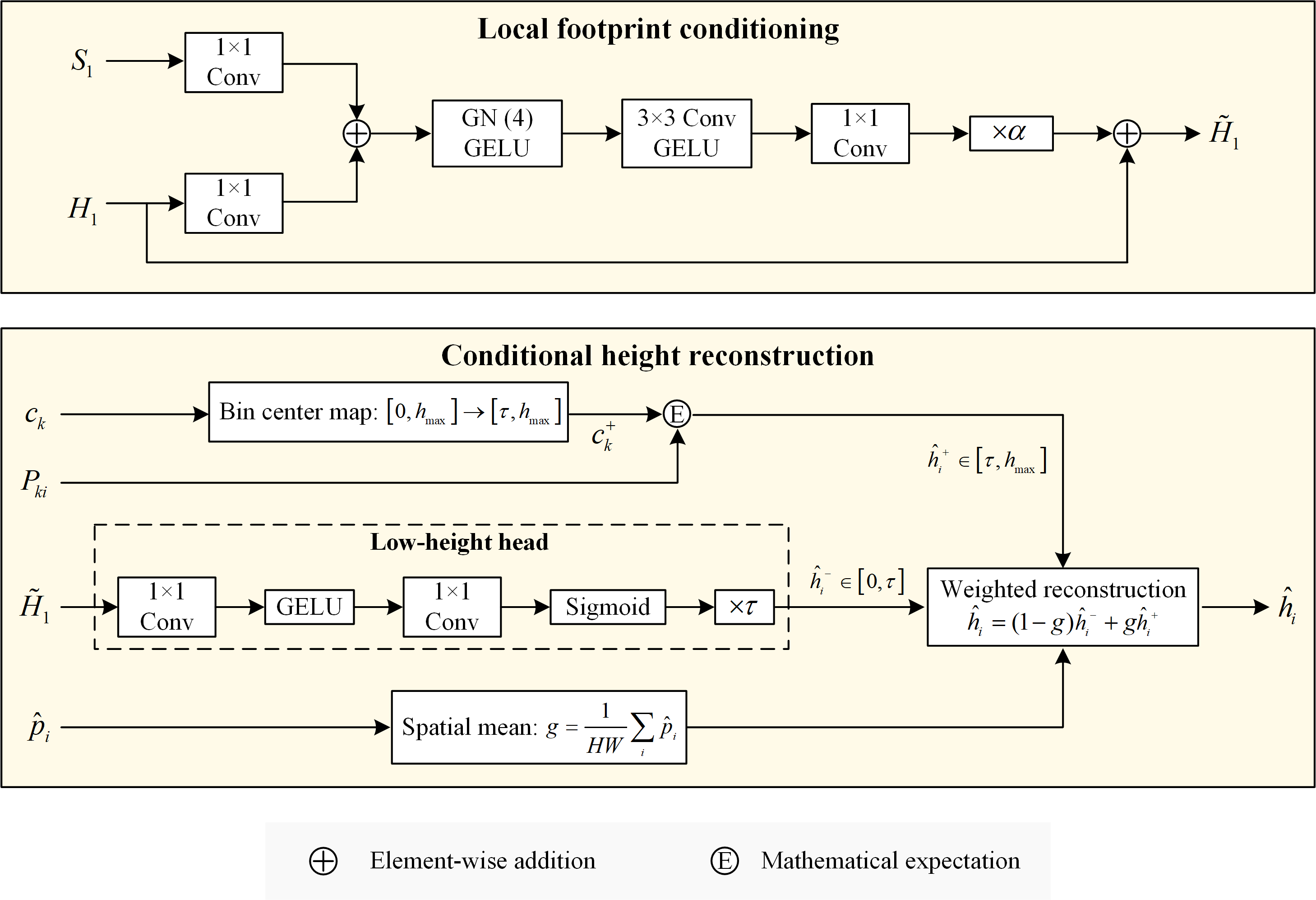}
\caption{\rev{The structure of FCHR (footprint-conditioned height reconstruction).}}
\label{fig:fchr}
\end{figure}

\rev{FCHR uses footprint information at two stages of height estimation, as shown in \figref{fig:fchr}.} \rev{It first conditions local height features and then combines complementary height estimates using a patch-shared weight derived from footprint probabilities.}

\rev{The full-resolution features $\mathbf H_1$ and $\mathbf S_1$ are projected to $D/4$ channels and combined by addition:}
{\color{RevisionText}
\begin{equation}
\label{eq:cross_task_context}
\begin{aligned}
\mathbf U&=\phi\!\left[\operatorname{GN}_4
(\mathcal P_h\mathbf H_1+\mathcal P_f\mathbf S_1)\right]\\
\widetilde{\mathbf H}_1&=\mathbf H_1+\alpha\,
\mathcal P_o\phi(\mathcal C_{3\times3}\mathbf U)
\end{aligned}
\end{equation}
}%
\noindent \rev{Here, $\phi$ denotes GELU and $\operatorname{GN}_4$ group normalization with four groups.} \rev{Pointwise projections $\mathcal P_h,\mathcal P_f$ map $D$ to $D/4$ channels; $\mathcal C_{3\times3}$ is a standard $D/4$-channel convolution, and $\mathcal P_o$ restores $D$ channels.} \rev{Zero initialization of $\mathcal P_o$ makes the conditioned feature initially equal to $\mathbf H_1$, while the fixed coefficient $\alpha=0.25$ scales the residual update.} \rev{The conditioned feature supplies both spatial assignment and the complementary reconstruction branch; footprint features and coarser height features remain unchanged.}

\rev{For height reconstruction, FCHR uses the footprint threshold $\tau=2\,$m to map the MDHR bin centers from $[0,h_{\max}]$ to $[\tau,h_{\max}]$ and computes their pixel-wise expectation:}
{\color{RevisionText}
\begin{equation}
\label{eq:conditional_height}
\begin{aligned}
c_k^{+}&=\tau+\frac{h_{\max}-\tau}{h_{\max}}c_k\\
\hat h_i^{+}&=\sum_{k=1}^{K}P_{ki}c_k^{+}
\end{aligned}
\end{equation}
}%
\noindent \rev{This order-preserving transformation confines the conditional estimate to the building-height range.} \rev{The low-height head $\mathcal C_l$ shown in \figref{fig:fchr} uses two $1\times1$ convolutions ($D\to D/2\to1$) with GELU between them, producing the complementary bounded estimate}
{\color{RevisionText}
\begin{equation}
\label{eq:low_height}
\hat{\mathbf h}^{-}=\tau\,\sigma\!\left[\mathcal C_l(\widetilde{\mathbf H}_1)\right]
\end{equation}
}%
\noindent \rev{Its output lies between 0 and $\tau$.} \rev{The final height combines the two estimates using the spatial mean of footprint probabilities:}
{\color{RevisionText}
\begin{equation}
\label{eq:gated_reconstruction}
\begin{aligned}
g&=\frac{1}{HW}\sum_{i=1}^{HW}\hat p_i\\
\hat h_i&=(1-g)\hat h_i^{-}+g\hat h_i^{+}
\end{aligned}
\end{equation}
}%
\noindent \rev{The patch-shared scalar $g$ summarizes estimated building support over the full grid, without reference labels or validity masking.} \rev{It weights continuous estimates rather than hard low/high classes; local variation remains in $\hat h_i^{-}$ and $\hat h_i^{+}$.} \rev{Height-loss gradients propagate through both estimates and footprint probabilities.}

\subsection{Multi-task objective}
\label{sec:objective}

Following existing studies that jointly model building height estimation and footprint segmentation ~\citep{liu2022associatively, cai2023deep, wang2024mf}, the training objective is formulated as a multi-task loss consisting of a height regression term and a footprint segmentation term. \rev{For image $b$ and pixel $i$, let $v_{bi}\in\{0,1\}$ indicate joint image--reference validity, and let $h_{bi}$ and $f_{bi}$ denote the reference height and its derived footprint target from Eq.~\eqref{eq:reference_footprint}.} \rev{The corresponding model outputs are the height estimate $\hat h_{bi}$ and the auxiliary footprint probability $\hat p_{bi}$; the additional index $b$ identifies the image in the mini-batch.} \rev{Using $\varepsilon=10^{-6}$ for numerical stability in both losses, the height objective is}
{\color{RevisionText}
\begin{equation}
\label{eq:height_loss}
\mathcal L_{\mathrm{MSE}}=
\frac{\sum_{b,i}v_{bi}(\hat h_{bi}-h_{bi})^2}{\sum_{b,i}v_{bi}+\varepsilon}
\end{equation}
}%
\noindent \rev{All valid pixels receive equal per-pixel weight.} \rev{The continuous reference heights are retained, including valid values below $\tau$; the threshold defines the auxiliary footprint target rather than replacing the regression target.}
\rev{Including non-building pixels in the height objective directly penalizes spurious height responses outside reference buildings and supports continuous height mapping over the valid image domain.}

\rev{For each image, the soft true-positive, false-negative, and false-positive counts are, respectively,}
{\color{RevisionText}
\begin{equation}
\label{eq:soft_counts}
\begin{aligned}
T_b&=\sum_i v_{bi}\hat p_{bi}f_{bi}\\
F_b^{-}&=\sum_i v_{bi}(1-\hat p_{bi})f_{bi}\\
F_b^{+}&=\sum_i v_{bi}\hat p_{bi}(1-f_{bi})
\end{aligned}
\end{equation}
}%
\noindent \rev{For a mini-batch of $N$ images, the footprint and joint objectives are}
{\color{RevisionText}
\begin{equation}
\label{eq:tversky_loss}
\mathcal L_{\mathrm{Tversky}}=1-\frac1N\sum_{b=1}^{N}
\frac{T_b+\varepsilon}{T_b+0.7F_b^{-}+0.3F_b^{+}+\varepsilon}
\end{equation}
}%
{\color{RevisionText}
\begin{equation}
\label{eq:total_loss}
\mathcal L=0.1\mathcal L_{\mathrm{MSE}}+\mathcal L_{\mathrm{Tversky}}
\end{equation}
}%
\noindent \rev{The larger false-negative coefficient emphasizes missed building support.} \rev{MSE is averaged over valid pixels in the mini-batch, whereas Tversky loss is calculated per image and then averaged across images.} \rev{The final height map and footprint probabilities receive direct supervision; bin generation, spatial assignment, and the reconstruction branches are optimized through the final height error, without separate bin-classification or branch-specific losses.}

\section{Experimental assessment}
\label{sec:experiments}

To assess PhiSat-2-based monocular BHE under the PHDataset setting and evaluate the proposed TSONet framework, comprehensive experiments are conducted from both quantitative and qualitative perspectives. \rev{First, comparisons with representative methods are performed to examine the overall performance of TSONet in building height estimation and height-map-derived building-region consistency.} Then, ablation studies are conducted to analyze the contributions of the proposed task-oriented modules and task configurations. \rev{In addition, sensor-information experiments and LiDAR-based accuracy auditing are introduced to assess the information value and practical relevance of PhiSat-2 imagery for BHE.} Together, these experiments evaluate how TSONet exploits PhiSat-2 spatial--spectral information under the current open-reference setting.\par

\subsection{\rev{Evaluation metrics and common reporting protocol}}
\label{sec:evaluation}

\rev{The pixel-level evaluation protocol assesses two aspects of the final height estimation: the accuracy of continuous building heights and the spatial consistency of building regions derived from the estimated height map.}
\rev{Comparisons, ablations, and PhiSat-2 input controls use the city-disjoint PHDataset described in Section~\ref{sec:dataset_construction}. The cross-sensor comparison uses the same test patch locations, whereas the LiDAR experiments use a separate matched subset or frozen-model audit as detailed in Section~\ref{sec:lidar_assessment}.}
All models are implemented in PyTorch. \rev{The main TSONet experiments use PyTorch 2.4.0 with CUDA 12.1 on a single NVIDIA Tesla V100 GPU with 16\,GB of memory.}
\rev{The repeated runs use random seeds 42--46, with metrics computed from pooled pixel-level error sums or confusion counts over the relevant evaluation subset for each run.} \rev{Results are reported as mean $\pm$ sample standard deviation across eligible runs, describing training variability on the fixed geographic partition.} \rev{Comparison methods retain their own optimization and initialization but share the input data, height-reference definition, and evaluation protocol.}

\begin{table}[t]
\centering\tableformat
\caption{Quantitative results of comparison experiments.}
\label{tab:comparison}
\begin{tabular}{@{}lccccc@{}}
\toprule
Method & \rev{$\mathrm{MAE}_b\downarrow$} & \rev{$\mathrm{RMSE}_b\downarrow$} & \rev{$\mathrm{RMSE}_{\mathrm{all}}\downarrow$} & IoU$\uparrow$ & Recall$\uparrow$ \\
\midrule
U-Net & \rev{\metriccell{9.461}{0.222}} & \rev{\metriccell{12.500}{0.158}} & \rev{\metriccell{5.190}{0.046}} & \rev{\metriccell{19.863}{3.177}}  & \rev{\metriccell{23.590}{4.417}} \\[2pt]
Eigen et al. & \rev{\metriccell{8.282}{0.234}} & \rev{\metriccell{11.638}{0.171}} & \rev{\metriccell{5.025}{0.013}} & \rev{\metriccell{26.238}{0.659}}  & \rev{\metriccell{69.790}{10.437}} \\[2pt]
Laina et al. & \rev{\metriccell{6.659}{0.246}} & \rev{\metriccell{10.121}{0.177}} & \rev{\metriccell{4.975}{0.078}} & \rev{\metriccell{31.985}{2.531}}  & \rev{\metriccell{79.497}{8.083}} \\[2pt]
BTS & \rev{\metriccell{6.117}{0.160}} & \rev{\metriccell{9.763}{0.134}} & \rev{\metriccell{4.921}{0.032}} & \rev{\metriccell{26.956}{1.292}}  & \rev{\metriccell{98.941}{0.426}} \\[2pt]
LocalBins & \rev{\bestmetriccell{5.893}{0.303}} & \rev{\bestmetriccell{9.464}{0.272}} & \rev{\metriccell{4.905}{0.062}} & \rev{\metriccell{28.610}{0.387}}  & \rev{\metriccell{98.953}{0.272}} \\[2pt]
DepthFormer & \rev{\metriccell{6.058}{0.141}} & \rev{\metriccell{9.584}{0.147}} & \rev{\metriccell{4.925}{0.044}} & \rev{\metriccell{26.664}{1.274}}  & \rev{\bestmetriccell{99.019}{0.428}} \\[2pt]
BinsFormer & \rev{\metriccell{5.995}{0.220}} & \rev{\metriccell{9.518}{0.231}} & \rev{\metriccell{4.901}{0.024}} & \rev{\metriccell{28.533}{0.711}}  & \rev{\metriccell{98.094}{0.577}} \\[2pt]
IM2HEIGHT & \rev{\metriccell{9.772}{0.237}} & \rev{\metriccell{12.819}{0.176}} & \rev{\metriccell{5.272}{0.042}} & \rev{\metriccell{16.708}{3.427}}  & \rev{\metriccell{19.441}{5.019}} \\[2pt]
DORN-height & \rev{\metriccell{9.400}{0.309}} & \rev{\metriccell{12.428}{0.226}} & \rev{\metriccell{5.233}{0.049}} & \rev{\metriccell{18.412}{3.980}}  & \rev{\metriccell{22.312}{6.033}} \\[2pt]
HTC-DC Net & \rev{\metriccell{6.868}{0.429}} & \rev{\metriccell{10.303}{0.407}} & \rev{\metriccell{5.069}{0.126}} & \rev{\metriccell{36.414}{0.703}}  & \rev{\metriccell{67.896}{7.861}} \\[2pt]
FusedSeg-HE & \rev{\metriccell{7.303}{0.290}} & \rev{\metriccell{10.541}{0.236}} & \rev{\bestmetriccell{4.696}{0.010}} & \rev{\bestmetriccell{36.611}{0.575}}  & \rev{\metriccell{67.285}{5.310}} \\[2pt]
TSONet & \rev{\metriccell{6.559}{0.108}} & \rev{\metriccell{9.928}{0.167}} & \rev{\metriccell{4.972}{0.108}} & \rev{\metriccell{34.640}{0.905}}  & \rev{\metriccell{84.245}{2.169}} \\[2pt]
\bottomrule
\end{tabular}
\end{table}

\Needspace{3\baselineskip}
{\predisplaypenalty=0
\rev{Negative estimated heights are clipped to zero before metric computation.} \rev{Height errors are not masked by the predicted footprint, so missed buildings remain part of the evaluation.}
\rev{Let $\Omega=\{i:v_i=1\}$ denote the valid evaluation pixels, $\Omega_h=\{i\in\Omega:h_i>\tau\}$ the reference building pixels, with $\tau=2\,$m as in Eq.~\eqref{eq:reference_footprint}. Height estimation is evaluated using building mean absolute error ($\mathrm{MAE}_b$), root mean squared error ($\mathrm{RMSE}_b$), and full-valid root mean squared error ($\mathrm{RMSE}_{\mathrm{all}}$):}
{\color{RevisionText}
\begin{equation}
\label{eq:evaluation_errors}
\begin{aligned}
\mathrm{MAE}_b&=\frac{1}{|\Omega_h|}\sum_{i\in\Omega_h}|\hat h_i-h_i|\\
\mathrm{RMSE}_b&=\sqrt{\frac{1}{|\Omega_h|}\sum_{i\in\Omega_h}(\hat h_i-h_i)^2}\\
\mathrm{RMSE}_{\mathrm{all}}&=\sqrt{\frac{1}{|\Omega|}\sum_{i\in\Omega}(\hat h_i-h_i)^2}
\end{aligned}
\end{equation}
}%
}
\noindent \rev{All reported height errors are in meters; IoU and Recall are expressed as percentages. A dash indicates an unavailable output. Parameter counts and estimated FLOPs are expressed in millions (M) and billions (G), respectively. Parameters are counted from the registered model, and FLOPs are estimated with the PyTorch profiler in evaluation mode for a $1\times7\times256\times256$ input, counting one multiply--accumulate as two operations. The estimates include profiler-supported operators; differences in operator coverage, including fused attention, preclude a complete computational-cost ranking, and no inference-latency comparison is claimed.} 

\begin{figure}[p]
\centering
\sbox{\comparisoncaptionbox}{\begin{minipage}{\textwidth}\caption{Qualitative results of comparison experiments.}\label{fig:compare}\end{minipage}}
\setlength{\comparisonheight}{\dimexpr\textheight-\ht\comparisoncaptionbox-\dp\comparisoncaptionbox-4pt\relax}
\includegraphics[width=\textwidth,height=\comparisonheight,keepaspectratio]{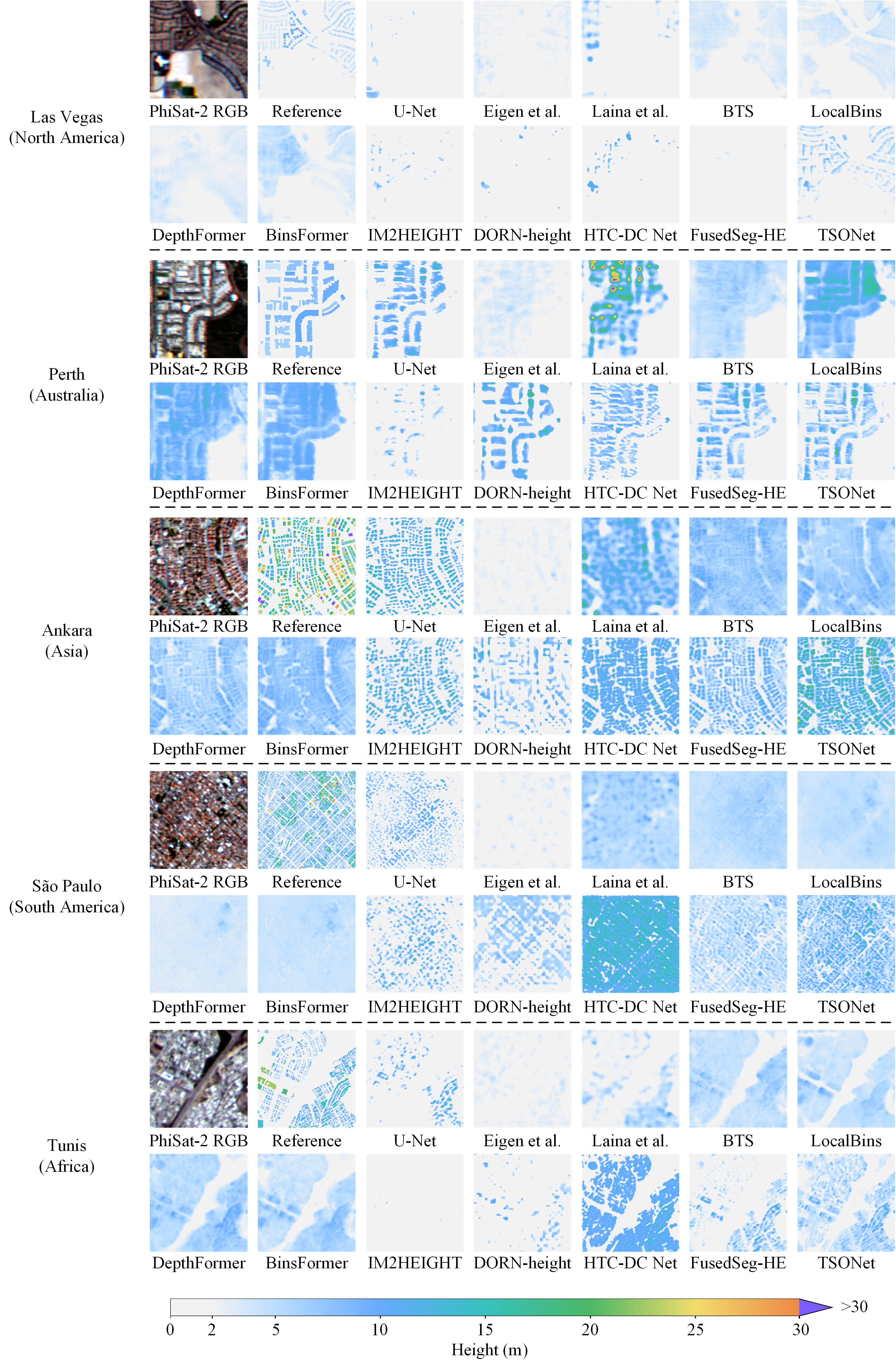}\par
\usebox{\comparisoncaptionbox}
\end{figure}

\rev{The principal spatial result is the footprint implied by the final height map.} \rev{IoU measures overall overlap, and Recall indicates retained reference-building support.} \rev{The height-derived mask $\hat f_i^{\,h}$ applies the threshold in Eq.~\eqref{eq:reference_footprint} to $\hat h_i$.} \rev{The auxiliary footprint head is evaluated separately:}
{\color{RevisionText}
\begin{equation}
\label{eq:evaluation_masks}
\hat f_i^{\,f}=\mathbf1(\hat p_i>0.5)
\end{equation}
}%
\noindent \rev{For either mask, true positives (TP), false positives (FP), and false negatives (FN) are accumulated against $f_i$ over $\Omega$.} \rev{The corresponding metrics are}
{\color{RevisionText}
\begin{equation}
\label{eq:evaluation_spatial}
\begin{aligned}
\mathrm{IoU}&=\frac{\mathrm{TP}}{\mathrm{TP}+\mathrm{FP}+\mathrm{FN}}\\
\mathrm{Recall}&=\frac{\mathrm{TP}}{\mathrm{TP}+\mathrm{FN}}
\end{aligned}
\end{equation}
}%
\indent \rev{The footprint head supplies auxiliary supervision and participates in height reconstruction; its scores are therefore presented as diagnostic results in the ablation analysis, whereas height-derived scores assess the spatial consistency of the principal output.}

\subsection{Comparison with representative methods}
\label{sec:comparison}

To comprehensively evaluate the performance of TSONet, three categories of competing methods are considered (\tabref{tab:comparison}). First, since TSONet is built upon a hierarchical encoder--decoder architecture analogous to U-Net~\citep{ronneberger2015u}, \rev{U-Net is included as a basic encoder--decoder comparison model}, using \rev{mean squared error (MSE) loss}. Second, given the strong similarity between MDE and MHE, several representative MDE methods are included for comparison, including Eigen et al.~\citep{eigen2014depth}, Laina et al.~\citep{laina2016deeper}, BTS~\citep{lee2019big}, LocalBins~\citep{bhat2022localbins}, DepthFormer~\citep{li2023depthformer}, and BinsFormer~\citep{li2024binsformer}. \rev{Finally, four representative MHE methods that are more directly related to the target task are also included, namely IM2HEIGHT~\citep{mou2018im2height}, DORN-height~\citep{li2020height}, HTC-DC Net~\citep{chen2023htc}, and FusedSeg-HE~\citep{gultekin2025fusing}.} \rev{The input interfaces of all methods are adapted to the seven PhiSat-2 bands.}

\label{sec:comparison_training}
\rev{TSONet is trained from scratch with the objective in Eq.~\eqref{eq:total_loss}, AdamW (peak learning rate $10^{-4}$; weight decay 0.01), and batch size 10. The schedule allows up to 30 epochs, with three warm-up epochs followed by cosine decay. Training stops after 10 epochs without improvement in validation building RMSE, and the checkpoint with the lowest validation value is tested; test results do not guide checkpoint or threshold selection. Comparison methods retain their original architectures and model-specific optimization, with inputs adapted to the seven PhiSat-2 bands. Pretrained encoders are used for Laina et al., DORN-height, HTC-DC Net, FusedSeg-HE, BTS, BinsFormer, and LocalBins; DepthFormer uses pretrained Swin-T weights. U-Net, Eigen et al., and IM2HEIGHT are trained from scratch. BinsFormer, BTS, DepthFormer, Eigen et al., and LocalBins additionally use a background regression term, MSE with coefficient 1.0, to constrain background height estimates not directly supervised by their native depth objectives. The comparison schedule also allows up to 30 epochs with an early-stopping patience of 10 epochs.}

\rev{For pretrained encoders, the first convolution or patch-projection kernel is expanded from three to seven input channels. The RGB kernels are reordered to match the B--G--R order of the first three PhiSat-2 bands; each remaining band is initialized with the mean of the RGB kernels, and all seven-channel weights are scaled by $3/7$. Other compatible encoder weights are retained, while task-specific layers are randomly initialized. For DepthFormer, this adaptation applies to the Swin-T branch; its shallow convolutional branch is initialized from scratch.}
\rev{No data augmentation was used in the comparison and ablation experiments.}

\tabref{tab:comparison} reports the quantitative comparison between TSONet and competing methods in terms of continuous height regression and height-map-derived building-region consistency. \rev{TSONet attains a building RMSE of 9.928\,m, MAE of 6.559\,m, and height-derived IoU of 34.640\%.} \rev{Relative to FusedSeg-HE and HTC-DC Net, which achieve the highest height-derived IoU among the MHE comparison methods, TSONet reduces building RMSE by 5.82\% and 3.64\%, respectively; its full-valid RMSE falls between those of the two methods. These methods attain higher IoU, whereas TSONet achieves higher Recall.} \rev{Conversely, MDE methods such as BinsFormer and LocalBins achieve lower building-height errors, whereas TSONet exceeds their height-derived IoU by 6.11 and 6.03 percentage points.} \rev{The lower building-height errors of these MDE methods therefore do not, by themselves, establish stronger performance on the joint task of height estimation and building-region delineation.}

\rev{\figref{fig:compare} compares selected scenes from the five PHDataset test cities: Las Vegas, Perth, Ankara, S\~ao Paulo, and Tunis.} \rev{Valid heights up to $2\,$m appear in gray, those between 2 and 30\,m use a continuous gradient, and those above $30\,$m appear in saturated purple.} \rev{Despite their lower building-height errors, LocalBins and BinsFormer produce smooth, spatially merged responses in the selected Ankara and S\~ao Paulo examples, making neighboring building regions difficult to distinguish. Together with their lower height-derived IoU in \tabref{tab:comparison}, these examples indicate limitations for spatially explicit building-height mapping.} \rev{Some MHE methods recover clear building patterns in individual scenes, including FusedSeg-HE in Perth and IM2HEIGHT in Ankara; TSONet retains recognizable building-layout cues across the selected cities.} \rev{Las Vegas and Perth have predominantly low reference heights, yet Laina et al. and LocalBins produce elevated responses in parts of these scenes.} \rev{Many height estimates have a compressed height range, particularly in Ankara and S\~ao Paulo. TSONet retains local height variation and spatial structure, although its height contrast remains below the reference.} \rev{In Tunis, height estimates are weak or diffuse, while TSONet preserves recognizable portions of the building layout.}

\begin{table}[t]
\centering\tableformat
\caption{\rev{Quantitative results of component and task ablation.}}
\label{tab:module_height}
\begin{tabular}{@{}lccccc@{}}
\toprule
\rev{Configuration} & \rev{$\mathrm{MAE}_b\downarrow$} & \rev{$\mathrm{RMSE}_b\downarrow$} & \rev{$\mathrm{RMSE}_{\mathrm{all}}\downarrow$} & IoU$\uparrow$ & Recall$\uparrow$ \\
\midrule
\rev{ST-baseline} & \rev{\metriccell{6.921}{0.430}} & \rev{\metriccell{10.293}{0.392}} & \rev{\metriccell{4.904}{0.107}} & \rev{\metriccell{33.715}{0.569}}  & \rev{\metriccell{78.816}{5.879}} \\[2pt]
\rev{MT-baseline} & \rev{\metriccell{6.743}{0.310}} & \rev{\metriccell{10.093}{0.295}} & \rev{\bestmetriccell{4.881}{0.085}} & \rev{\metriccell{34.243}{0.814}}  & \rev{\metriccell{80.099}{2.670}} \\[2pt]
\rev{MT + FCHR} & \rev{\metriccell{6.614}{0.208}} & \rev{\metriccell{10.040}{0.203}} & \rev{\metriccell{4.889}{0.050}} & \rev{\metriccell{34.037}{0.720}}  & \rev{\bestmetriccell{84.446}{2.391}} \\[2pt]
\rev{MT + MDHR} & \rev{\metriccell{6.749}{0.388}} & \rev{\metriccell{10.141}{0.384}} & \rev{\metriccell{4.897}{0.133}} & \rev{\metriccell{33.598}{0.995}}  & \rev{\metriccell{82.434}{4.292}} \\[2pt]
TSONet & \rev{\bestmetriccell{6.559}{0.108}} & \rev{\bestmetriccell{9.928}{0.167}} & \rev{\metriccell{4.972}{0.108}} & \rev{\bestmetriccell{34.640}{0.905}}  & \rev{\metriccell{84.245}{2.169}} \\[2pt]
\bottomrule
\end{tabular}
\par\vspace{12pt}

\centering\tableformat\setlength{\tabcolsep}{12pt}
\caption{\rev{Independent footprint-head diagnostics for the ablation.}}
\label{tab:module_fp}
\begin{tabular}{@{}lcc@{}}
\toprule
\rev{Configuration} & IoU$\uparrow$ & Recall$\uparrow$ \\
\midrule
\rev{ST-baseline} & \rev{--}  &  -- \\[2pt]
\rev{MT-baseline} & \rev{\metriccell{33.705}{1.320}}  & \rev{\metriccell{88.891}{3.096}} \\[2pt]
\rev{MT + FCHR} & \rev{\metriccell{33.532}{0.893}}  & \rev{\metriccell{88.166}{3.091}} \\[2pt]
\rev{MT + MDHR} & \rev{\metriccell{33.474}{1.011}}  & \rev{\metriccell{90.497}{1.283}} \\[2pt]
\rev{TSONet} & \rev{\metriccell{34.358}{1.167}}  & \rev{\metriccell{87.453}{2.213}} \\[2pt]
\bottomrule
\end{tabular}
\end{table}

\begin{figure}[t]
\centering
\includegraphics[width=\textwidth,height=0.78\textheight,keepaspectratio]{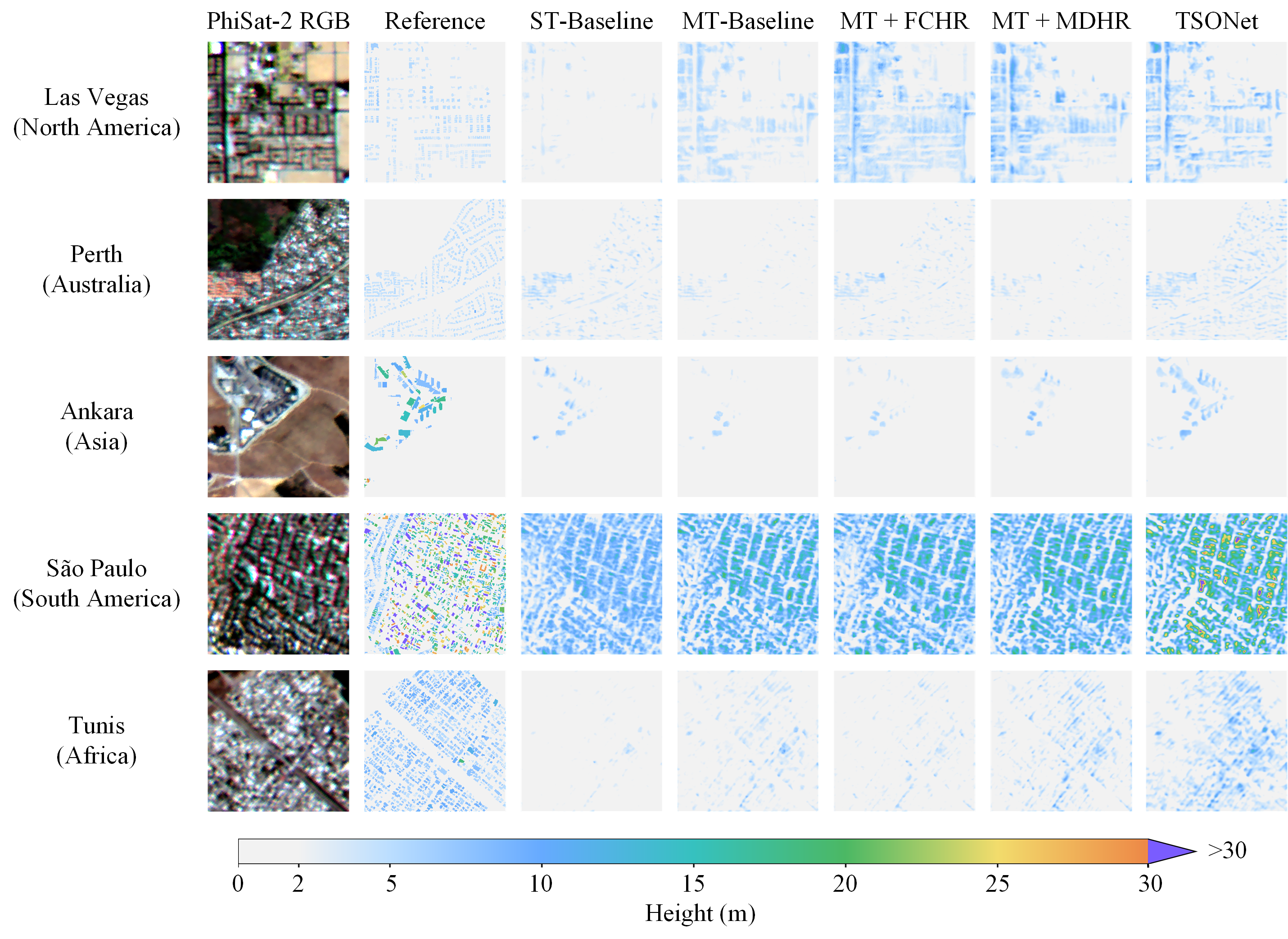}
\caption{\rev{Qualitative results of component and task ablation.}}
\label{fig:ablation}
\end{figure}

\subsection{\rev{Component and task ablation}}
\label{sec:module_ablation}
\label{sec:task_ablation}

\rev{The ablation configurations use the TSONet settings described in Section~\ref{sec:comparison}. Five configurations are compared: a height-only single-task baseline (ST-baseline), a multitask baseline (MT-baseline), MT + FCHR, MT + MDHR, and the complete TSONet. The shared baseline height pathway uses a BinsFormer-style structure~\citep{li2024binsformer}: updated queries generate adaptive bin widths and query embeddings, and query--pixel dot products produce spatial assignments. Height is reconstructed as the expectation over bins spanning 0--145\,m. ST-baseline omits the footprint decoder and head and uses $0.1\mathcal L_{\mathrm{MSE}}$ over all valid pixels. MT-baseline adds the auxiliary footprint pathway and Tversky supervision, but includes neither of the two FCHR stages described in Section~\ref{sec:fchr}.}

\rev{MDHR replaces the baseline height representation with the multiscale query and convolutional-assignment pathways in Section~\ref{sec:mdhr}.} \rev{Without FCHR, these pathways use the unconditioned height feature and reconstruct height directly over 0--145\,m, without local cross-task conditioning or footprint-weighted reconstruction.} \rev{FCHR alone retains the baseline query and assignment mechanisms, adds local feature conditioning, and combines conditional-bin and low-height estimates using a footprint-derived weight (Section~\ref{sec:fchr}).} \rev{The complete model combines the two components as defined in Section~\ref{sec:method}.} \rev{Thus, the ablations assess architectural and task contributions under fixed loss coefficients and full-valid height supervision. TSONet has 43.104 million registered parameters and an estimated 202.684\,GFLOPs per $1\times7\times256\times256$ input. Relative to MT-baseline, the full TSONet has 2.33\% fewer registered parameters but 8.45\% higher estimated FLOPs (\tabref{tab:model_complexity}).}

\rev{Relative to ST-baseline, the complete model reduces building RMSE and MAE by 3.54\% and 5.22\%, respectively, increases height-derived IoU by 0.925 percentage points, and raises Recall from 78.816\% to 84.245\%.} \rev{Relative to MT-baseline, the respective height-error reductions are 1.63\% and 2.72\%, accompanied by a small increase in IoU.} \rev{Including both baselines distinguishes the contribution of the complete architecture from the effect of auxiliary supervision alone.} \rev{FCHR reduces the mean building RMSE of MT-baseline from 10.093 to 10.040\,m and MAE from 6.743 to 6.614\,m.} \rev{Adding MDHR to this conditioned reconstruction further reduces these errors to 9.928 and 6.559\,m, while height-derived IoU increases from 34.037\% to 34.640\%.} \rev{MDHR alone does not reduce the mean building error relative to MT-baseline.} \rev{Its contribution is therefore a modest increment within the complete conditioned design, rather than a uniform independent improvement.}

\begin{table}[t]
\centering\tableformat\setlength{\tabcolsep}{10pt}
\caption{\rev{Parameter counts and estimated inference FLOPs of the comparison and ablation models.}}
\label{tab:model_complexity}
\begin{tabular}{@{}lcc@{}}
\toprule
\rev{Model or configuration} & \rev{Params (M)$\downarrow$} & \rev{Estimated FLOPs (G)$\downarrow$}  \\
\midrule
\multicolumn{3}{@{}l@{}}{\rev{\textbf{Comparison models}}} \\
U-Net & \rev{13.394} & \rev{62.143}  \\[2pt]
Eigen et al. & \rev{71.067} & \rev{4.920}  \\[2pt]
Laina et al. & \rev{63.584} & \rev{29.631}  \\[2pt]
BTS & \rev{66.487} & \rev{56.655}  \\[2pt]
LocalBins & \rev{73.707} & \rev{38.185}  \\[2pt]
DepthFormer & \rev{52.715} & \rev{86.745}  \\[2pt]
BinsFormer & \rev{43.927} & \rev{37.245}  \\[2pt]
\rev{IM2HEIGHT} & \rev{7.481} & \rev{55.303}  \\[2pt]
DORN-height & \rev{74.431} & \rev{164.822}  \\[2pt]
HTC-DC Net & \rev{56.853} & \rev{113.451}  \\[2pt]
FusedSeg-HE & \rev{125.709} & \rev{73.493}  \\[2pt]
TSONet & \rev{43.104} & \rev{202.684}  \\[2pt]
\midrule
\multicolumn{3}{@{}l@{}}{\rev{\textbf{Ablation configurations}}} \\
\rev{ST-baseline} & \rev{33.423} & \rev{114.076}  \\[2pt]
\rev{MT-baseline} & \rev{44.132} & \rev{186.897}  \\[2pt]
\rev{MT + FCHR} & \rev{44.140} & \rev{187.884}  \\[2pt]
\rev{MT + MDHR} & \rev{42.877} & \rev{201.697}  \\[2pt]
\bottomrule
\end{tabular}
\end{table}

\rev{\tabref{tab:module_fp} reports the auxiliary footprint head separately.} \rev{ST-baseline has no footprint head and therefore no score for this output.} \rev{TSONet attains a mean footprint-head IoU of 34.358\%, compared with 33.705\% for MT-baseline.} \rev{This diagnostic complements the principal height-derived spatial scores rather than replacing them. The comparable footprint-head and height-derived IoUs of the full model (34.358\% and 34.640\%, respectively) indicate similar aggregate overlap performance against the shared reference footprint.}

\rev{\figref{fig:ablation} compares the task and component configurations in five test cities.} \rev{Multitask configurations recover clearer building responses than ST-baseline in scenes such as Las Vegas and Tunis, although their relative performance varies across cities. The full TSONet preserves recognizable building-layout patterns in all selected scenes.} \rev{In S\~ao Paulo, TSONet exhibits greater height contrast than the other configurations, including height estimates above 25\,m, although the reference retains finer building separation and more high-valued structures.} \rev{The Perth and Tunis scenes show that individual small buildings remain difficult to resolve, while all configurations underestimate several reference heights in the Ankara scene.} \rev{The effects of MDHR and FCHR vary across these patches rather than yielding a uniform visual ordering, consistent with their modest aggregate differences in \tabref{tab:module_height}.}

\begin{table}[t]
\centering\tableformat
\caption{\rev{Sensor comparison and PhiSat-2 channel configurations using TSONet.}}
\label{tab:sensor}
\label{tab:input_controls}
\begin{tabular*}{\textwidth}{@{\extracolsep{\fill}}lccccc@{}}
\toprule
\rev{Input} & \rev{$\mathrm{MAE}_b\downarrow$} & \rev{$\mathrm{RMSE}_b\downarrow$} & \rev{$\mathrm{RMSE}_{\mathrm{all}}\downarrow$} & IoU$\uparrow$ & Recall$\uparrow$ \\
\midrule
\rev{PhiSat-2 (7 bands)} & \rev{\metriccell{6.559}{0.108}} & \rev{\metriccell{9.928}{0.167}} & \rev{\metriccell{4.972}{0.108}} & \rev{\metriccell{34.640}{0.905}}  & \rev{\metriccell{84.245}{2.169}} \\[2pt]
\rev{PhiSat-2 (RGB)} & \rev{\metriccell{6.976}{0.141}} & \rev{\metriccell{10.424}{0.131}} & \rev{\metriccell{4.894}{0.047}} & \rev{\metriccell{34.915}{0.632}}  & \rev{\metriccell{79.268}{2.934}} \\[2pt]
\rev{PhiSat-2 (RGB + NIR)} & \rev{\metriccell{6.968}{0.301}} & \rev{\metriccell{10.353}{0.303}} & \rev{\metriccell{4.942}{0.081}} & \rev{\metriccell{34.322}{0.602}}  & \rev{\metriccell{79.007}{4.853}} \\[2pt]
\rev{Sentinel-2 (7 bands)} & \rev{\metriccell{6.660}{0.288}} & \rev{\metriccell{10.313}{0.266}} & \rev{\metriccell{5.058}{0.135}} & \rev{\metriccell{34.338}{1.623}}  & \rev{\metriccell{81.038}{4.594}} \\[2pt]
\rev{Landsat 8 (4 bands)} & \rev{\metriccell{6.956}{0.210}} & \rev{\metriccell{10.568}{0.237}} & \rev{\metriccell{5.093}{0.088}} & \rev{\metriccell{29.554}{0.569}}  & \rev{\metriccell{79.447}{3.426}} \\[2pt]
\bottomrule
\end{tabular*}
\par\vspace{12pt}
\centering
\includegraphics[width=\textwidth,height=0.72\textheight,keepaspectratio]{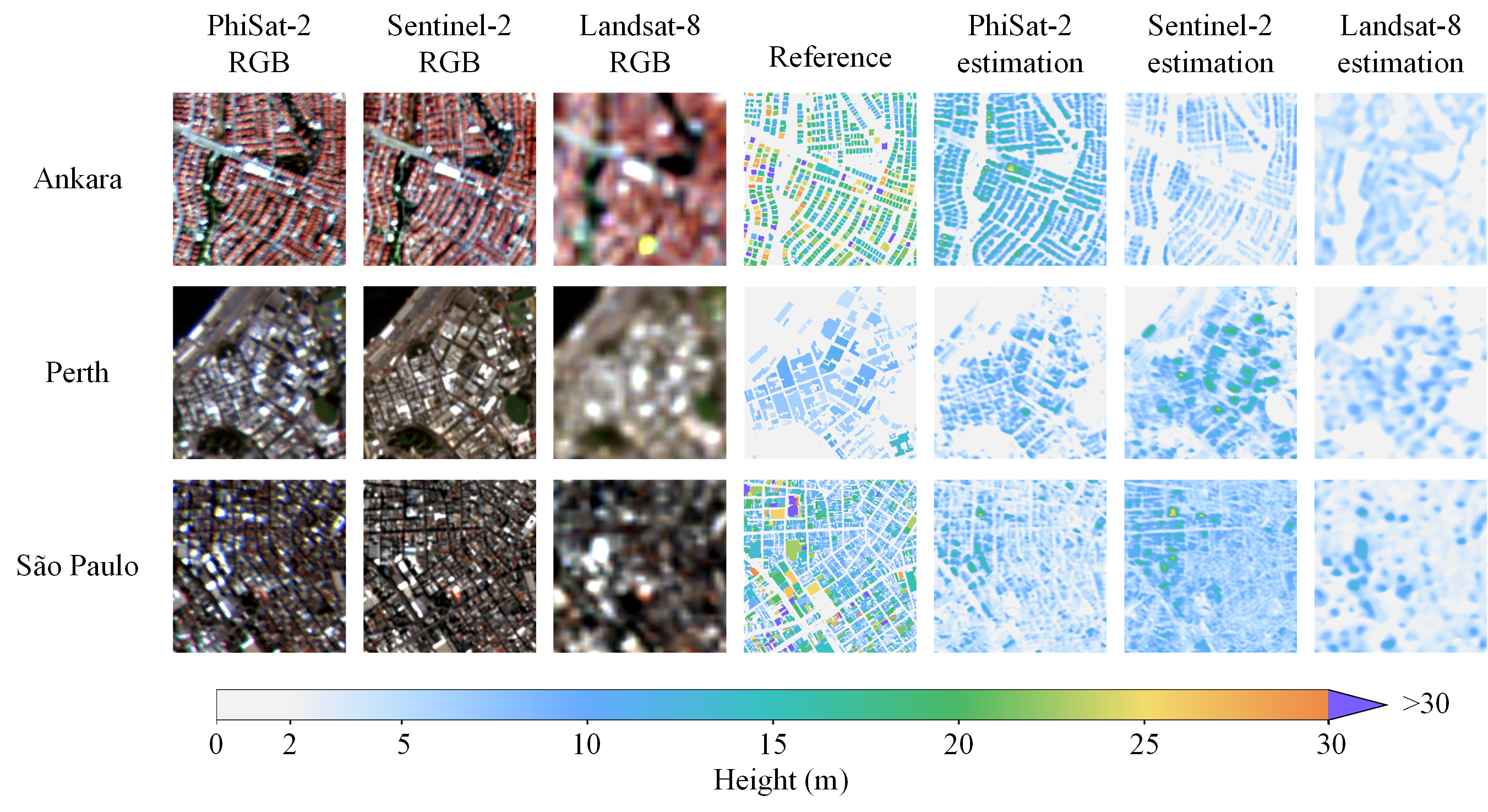}
\captionof{figure}{\rev{Qualitative comparison of building-height estimation using PhiSat-2, Sentinel-2, and Landsat 8 imagery.}}
\label{fig:compare_sensors}

\end{table}

\subsection{\rev{Sensor-information assessment of PhiSat-2 imagery}}
\label{sec:sensor_info}

\rev{To assess the utility of PhiSat-2 imagery for building-height estimation, TSONet is evaluated with 7-band PhiSat-2, 7-band Sentinel-2, and 4-band Landsat 8 inputs, and retrained with PhiSat-2 RGB or RGB + NIR inputs to examine the contribution of spectral channels. The 7-band PhiSat-2 reference is the main TSONet configuration. Band selection for the other sensors seeks broad spectral correspondence with PhiSat-2: Sentinel-2 uses visible bands B2--B4, red-edge bands B5--B7, and near-infrared band B8, whereas Landsat 8 uses visible bands B2--B4 and near-infrared band B5 because it lacks corresponding red-edge bands. These configurations cover similar spectral regions but do not have identical spectral responses.}

\rev{All configurations use the main city-disjoint patch locations, the same $256\times256$ grid and ground footprint, and the same height references. Sentinel-2 and Landsat 8 are aligned by geospatial bilinear interpolation, which does not increase their native spatial resolution; their stored values are multiplied by $10^{-4}$ to recover top-of-atmosphere reflectance. The Sentinel-2 and Landsat 8 experiments use the same TSONet architecture, objective, training schedule, and validation-selection rule as the main PhiSat-2 experiment; only the input data source and its associated channel interface and preprocessing are adapted. The PhiSat-2 channel controls also retain the original 7-channel validity mask and input preprocessing. Preprocessing differs between sensors, so their comparison assesses the available configurations rather than isolating sensor effects alone.}

\rev{As shown in \tabref{tab:sensor}, the 7-band PhiSat-2 configuration yields a mean building RMSE of 9.928\,m, compared with 10.313\,m for Sentinel-2 and 10.568\,m for Landsat 8, corresponding to differences of 3.73\% and 6.05\%, respectively. Its height-derived IoU is higher by 0.302 and 5.086 percentage points, respectively, and its building MAE is also lower than those of both sensors. Although the differences relative to Sentinel-2 are modest, particularly in MAE and IoU, the 7-band PhiSat-2 configuration leads on all reported mean metrics relative to Sentinel-2 and Landsat 8.}

\rev{Complementing these quantitative results, \figref{fig:compare_sensors} shows clearer building groups and boundaries with PhiSat-2 than with Landsat 8, whose estimates merge neighboring structures into diffuse responses. Improvements over Sentinel-2 are more modest and scene-dependent: PhiSat-2 better preserves height contrast in Ankara and reduces height overestimation in Perth. These examples highlight local benefits for both building-region delineation and height estimation.}

\rev{The within-PhiSat-2 comparisons further examine the contribution of spectral channels. Using all 7 channels lowers mean building RMSE by 4.76\% relative to RGB and 4.10\% relative to RGB + NIR, indicating a benefit from the fuller spectral input for building-height estimation. This benefit does not extend uniformly to all metrics: RGB inputs yield a slightly higher height-derived IoU and lower full-valid RMSE, but lower recall than the 7-band configuration. Building-height accuracy and the spatial consistency of the height map therefore respond differently to channel availability.}

\rev{Even with fewer bands, PhiSat-2 retains advantages in some metrics over the other sensors. Relative to Landsat 8, its RGB and RGB + NIR inputs lower mean building RMSE by 1.36\% and 2.03\%, respectively, and increase IoU by 5.361 and 4.768 percentage points, although neither improves building MAE or recall. PhiSat-2 RGB + NIR and 7-band Sentinel-2 have close mean building RMSE values (10.353 versus 10.313\,m) and IoU values (34.322\% versus 34.338\%), while Sentinel-2 retains lower building MAE and higher recall. Both reduced-band PhiSat-2 configurations also have lower full-valid RMSE than Sentinel-2 and Landsat 8. Taken together, these comparisons support the utility of PhiSat-2 across the tested spectral configurations, with all 7 bands providing the lowest building errors.}

\begin{table}[t]
\centering\tableformat
\caption{\rev{LiDAR-based assessment of supervision sources and frozen models.}}
\label{tab:lidar_assessment}
\label{tab:lidar_supervision}
\label{tab:lidar_common}
\begin{tabular*}{\textwidth}{@{\extracolsep{\fill}}lcc@{}}
\toprule
\multicolumn{3}{@{}l@{}}{\rev{\textbf{(a) Matched supervision-source experiment}}} \\
\rev{Supervision source} & \rev{$\mathrm{MAE}_b\downarrow$} & \rev{$\mathrm{RMSE}_b\downarrow$} \\
\midrule
\rev{Open-derived} & \rev{\metriccell{4.234}{0.247}} & \rev{\metriccell{6.751}{0.317}}  \\[2pt]
\rev{LiDAR-derived} & \rev{\metriccell{3.685}{0.132}} & \rev{\metriccell{5.868}{0.223}}  \\[2pt]
\end{tabular*}
\par\medskip
\begin{tabular*}{\textwidth}{@{\extracolsep{\fill}}lccccc@{}}
\multicolumn{6}{@{}l@{}}{\rev{\textbf{(b) Frozen main models}}} \\
\rev{Model} & \rev{\shortstack{\\$\mathrm{MAE}_b\downarrow$}} & \rev{\shortstack{\\$\mathrm{RMSE}_b\downarrow$}} & \rev{\shortstack{Barcelona\\$\mathrm{RMSE}_b\downarrow$}} & \rev{\shortstack{Madrid\\$\mathrm{RMSE}_b\downarrow$}} & \rev{\shortstack{London\\$\mathrm{RMSE}_b\downarrow$}} \\
\midrule
BinsFormer & \rev{\metriccell{5.904}{0.190}} & \rev{\metriccell{8.470}{0.231}} & \rev{\metriccell{9.025}{0.259}} & \rev{\metriccell{9.451}{0.336}} & \rev{\metriccell{6.746}{0.168}}  \\[2pt]
HTC-DC Net & \rev{\metriccell{7.161}{0.360}} & \rev{\metriccell{9.517}{0.425}} & \rev{\metriccell{9.066}{0.504}} & \rev{\metriccell{10.286}{0.655}} & \rev{\metriccell{9.092}{0.262}}  \\[2pt]
TSONet & \rev{\metriccell{6.902}{0.320}} & \rev{\metriccell{9.211}{0.380}} & \rev{\metriccell{9.075}{0.297}} & \rev{\metriccell{9.802}{0.686}} & \rev{\metriccell{8.698}{0.342}}  \\[2pt]
\bottomrule
\end{tabular*}
\end{table}

\subsection{\rev{LiDAR-based assessment}}
\label{sec:lidar_assessment}

\rev{LiDAR is used to compare two training-supervision sources for TSONet (\tabref{tab:lidar_assessment}(a)) and to assess models already trained on PHDataset without retraining (\tabref{tab:lidar_assessment}(b)). Both assessments are limited to building-height errors because background coverage is incomplete. Building support comes from map-derived footprints in London and the LiDAR building class in Barcelona and Madrid. Eligible pixels have finite LiDAR heights above $2\,$m, at least 50\% valid source coverage, and consistent heights in overlapping source tiles.}

\rev{The supervision comparison uses identical imagery: 855 training and 197 spatially separated validation patches from Barcelona and Madrid, and 495 test patches from London. Both variants share normalization parameters fitted on these 855 training patches and applied unchanged to all three subsets. Open-derived training labels follow the PHDataset approach, whereas LiDAR-derived training labels use LiDAR heights; each source also defines its auxiliary footprint targets through the $2\,$m height threshold. Both variants use LiDAR heights for validation-based model selection and final testing. Architecture and training settings are held fixed, with matched initialization across paired runs, so the comparison isolates the change in joint height and footprint supervision.}

\rev{LiDAR-derived supervision reduces London building RMSE from 6.751 to 5.868\,m and MAE from 4.234 to 3.685\,m, decreases of 13.08\% and 12.96\%, respectively. With architecture held fixed, this improvement identifies joint supervision source as a contributor to TSONet's remaining error. The open-derived result also supports PHDataset-style supervision for a model evaluated against LiDAR in a held-out city, with LiDAR-based validation used for model selection.}

\rev{For the frozen-model experiment in \tabref{tab:lidar_assessment}(b), the original main-model checkpoints of TSONet, BinsFormer, and HTC-DC Net are evaluated without retraining or LiDAR-based checkpoint selection. BinsFormer and HTC-DC Net represent adaptive-bin depth estimation and building-height estimation with auxiliary spatial information, respectively. All models use the same 1,569 patches from Barcelona, Madrid, and London. In PHDataset, Barcelona and Madrid belong to the main training partition and London to validation, so this experiment provides an alternative-reference assessment rather than an unseen-city generalization test. The overall metrics are computed from pixel-level errors across the three cities; the remaining columns give city-specific building RMSE.}

\rev{Under the common LiDAR reference, TSONet achieves an overall building RMSE of 9.211\,m, compared with 9.517\,m for HTC-DC Net and 8.470\,m for BinsFormer. This ranking agrees with \tabref{tab:comparison}, supporting TSONet's performance relative to HTC-DC Net under a different reference. Its gap to BinsFormer is smaller in Barcelona (9.075 versus 9.025\,m) than in London (8.698 versus 6.746\,m). Although BinsFormer achieves lower height errors, its lower height-derived building-region IoU in the main comparison highlights the trade-off between height accuracy and spatial consistency, supporting their joint assessment in PHDataset.}

\rev{Overall, these experiments support PHDataset-style supervision under the tested training and LiDAR-based model-selection conditions, and assess frozen models against an alternative reference in the main training and validation cities. Their geographic and reference-source coverage does not establish the quality of all PHDataset labels or the underlying public products.}

\subsection{Contextual comparison with global building height products}
\label{sec:product_context}
\begin{figure*}[t]
\centering
\includegraphics[width=\textwidth]{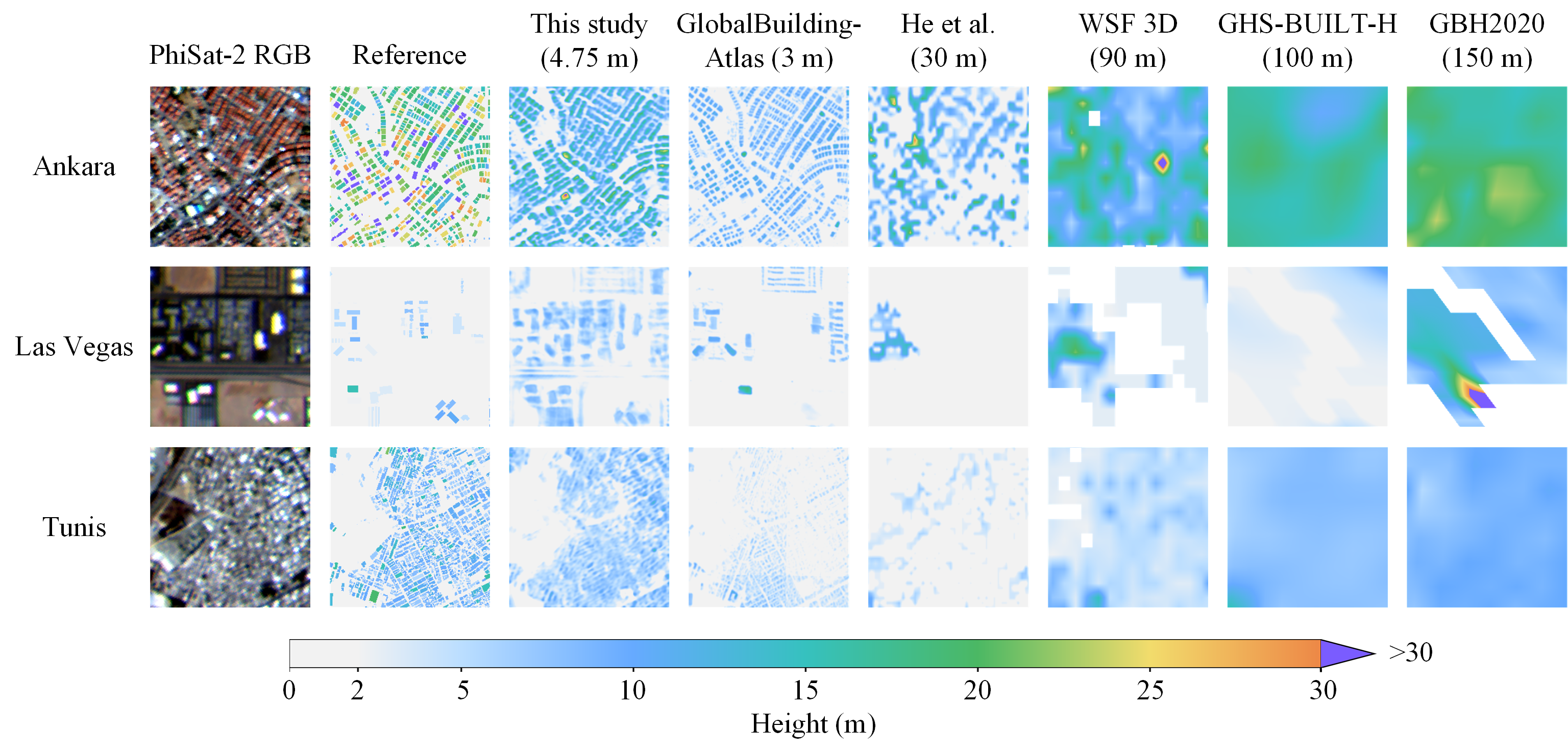}
\caption{Patch-level qualitative comparison between this study and global building height products. \rev{White gaps in the WSF 3D and GBH2020 panels represent missing source values rendered transparent against a white background.}}
\label{fig:product_comparison}
\end{figure*}

To further place PhiSat-2-based BHE in the context of existing global height products, a patch-level qualitative comparison was conducted against several publicly available global building height products with different spatial resolutions, including GlobalBuildingAtlas~\citep{zhu2025globalbuildingatlas} at 3\,m, He et al.~\citep{he2023global} at 30\,m, WSF 3D~\citep{esch2022world} at 90\,m, GHS-BUILT-H~\citep{Pesaresi2023GHSBuiltH} at 100\,m, and GBH2020~\citep{ma2023globalproductfinescaleurban} at 150\,m. \rev{Selected test patches from Ankara, Las Vegas, and Tunis are shown in \figref{fig:product_comparison}.} It should be noted that these products differ in spatial resolution, spatial reference, and product characteristics. Therefore, for patch-level visual comparison, all products were first projected to the WGS 1984 Web Mercator projection (EPSG:3857) and then resampled onto the 4.75\,m patch grid using bilinear interpolation, so that all results could be displayed over the same spatial extent and grid layout.

\rev{Compared with coarser global products, which mainly reflect broader neighborhood- or block-scale height patterns, both the TSONet estimates and GlobalBuildingAtlas provide more detailed local height variations and building-level structures in the selected patches.} GlobalBuildingAtlas generally presents finer boundary details, which is consistent with its higher nominal spatial resolution. \rev{In Ankara, TSONet retains the dense, regularly arranged building pattern and stronger height contrasts visible in the reference, whereas GlobalBuildingAtlas depicts sharper individual building outlines but generally lower heights.} \rev{In Las Vegas, TSONet recovers more building responses than GlobalBuildingAtlas, although some responses extend beyond the reference building regions.} \rev{In Tunis, TSONet preserves a more continuous response over the densely built area, while many GlobalBuildingAtlas heights remain close to the lower end of the displayed scale.} \rev{This qualitative comparison illustrates the potential of PhiSat-2-based BHE for mapping relatively detailed urban vertical structure, while providing product-level context rather than strict ground-truth validation.}

\section{Discussion}
\label{sec:discussion}
\rev{Understanding PhiSat-2-based building-height estimation requires examining both the composition of the reference data and the geometric consistency of multispectral observations.} \rev{The relationships between city- and source-specific height distributions and estimation errors are therefore examined, together with the effects of inter-band misalignment on building-region delineation.}

\Needspace{5\baselineskip}
\subsection{\rev{Performance variation across cities and height distributions}}
\label{sec:source_stratified}
\label{sec:city_results}

\begin{figure}[t]
\centering
\includegraphics[width=0.95\textwidth]{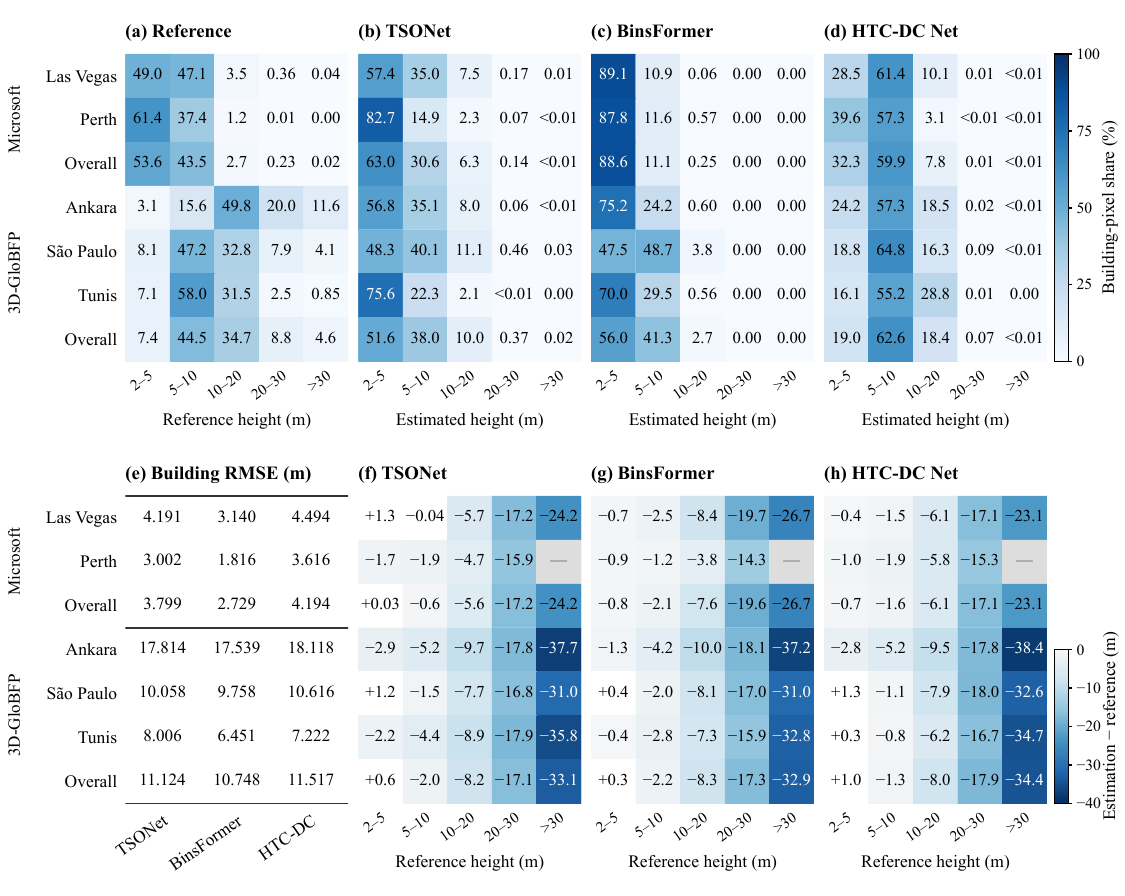}
\caption{\rev{Reference and estimated height distributions and signed biases for representative methods: (a) shared reference distribution; (b--d) estimated distributions for TSONet, BinsFormer, and HTC-DC Net; (e) building RMSE; and (f--h) bias stratified by reference height. Each Overall row pools the cities supported by the corresponding reference source. Distributions are normalized over valid reference heights $>2\,$m in (a) and valid predictions $>2\,$m in (b--d); (e--h) retain all valid reference-building pixels. Model values are averaged across seeds 42--46 after within-city/source pooling. Bias is estimated minus reference height. Blue shading indicates negative bias; positive biases retain their signed values on a white background. Gray cells indicate empty reference intervals. Finite intervals are left-open and right-closed.}}
\label{fig:city_height_error}
\end{figure}

\rev{\figref{fig:city_height_error} compares TSONet with BinsFormer and HTC-DC Net using the same test subset and trained models as \tabref{tab:comparison}.} \rev{BinsFormer and HTC-DC Net are selected as representative MDE and MHE methods, respectively, because of their strong building-height estimation performance.} \rev{The test set contains 354 patches from Las Vegas, 228 from Perth, 394 from Ankara, 506 from S\~ao Paulo, and 238 from Tunis (\tabref{tab:dataset_inventory}).} \rev{All three models have lower building RMSE in Microsoft-supported Las Vegas and Perth than in the three 3D-GloBFP-supported cities, with Perth lowest and Ankara highest (\figref{fig:city_height_error}e).} \rev{The source-level Overall rows show the same contrast: building RMSE is 3.799 versus 11.124\,m for TSONet, 2.729 versus 10.748\,m for BinsFormer, and 4.194 versus 11.517\,m for HTC-DC Net on the Microsoft and 3D-GloBFP subsets, respectively.} \rev{The reference distributions differ accordingly: the $(10,20]$\,m interval accounts for 2.7\% of Microsoft-supported building pixels and 34.7\% of 3D-GloBFP-supported pixels, with larger shares in the higher intervals for the latter source (\figref{fig:city_height_error}a).} \rev{3D-GloBFP offers a practical building-level reference where alternative height data are limited.} \rev{Its original evaluation reported higher correlation and lower RMSE than the Microsoft Global ML Building Footprints heights assessed in six US cities, although that comparison used an earlier Microsoft data snapshot and does not establish a ranking for the study areas considered here~\citep{che20243d}.}

\rev{The estimated distributions are concentrated in lower height intervals and contain far fewer tall-building responses than the reference distribution (\figref{fig:city_height_error}a--d).} \rev{For example, in Ankara, the $(20,30]$\,m shares are 20.0\% for the reference and 0.06\%, 0.00\%, and 0.02\% for the TSONet, BinsFormer, and HTC-DC Net distributions, respectively, each normalized over its valid predictions above 2\,m.} \rev{The increasingly negative biases at greater reference heights provide complementary evidence of tall-building underestimation in all three models (\figref{fig:city_height_error}f--h).}

\rev{Together, these distributions and biases suggest that height composition contributes to accuracy differences across cities and reference-source groups.} \rev{However, biases also differ within broad reference-height intervals: all three models show a larger negative bias in Ankara than in Perth within $(5,10]$\,m.} \rev{Urban morphology, geographic transfer, and spatial or temporal image--reference mismatches may also contribute.} \rev{Reducing tall-building underestimation and improving cross-city consistency are therefore shared priorities for future model and reference-data development.}

\Needspace{3\baselineskip}
\subsection{Sensitivity to geometric inconsistencies}
\label{sec:geometric_sensitivity}
\begin{figure}[t]
\centering
\includegraphics[width=0.95\textwidth,height=0.38\textheight,keepaspectratio]{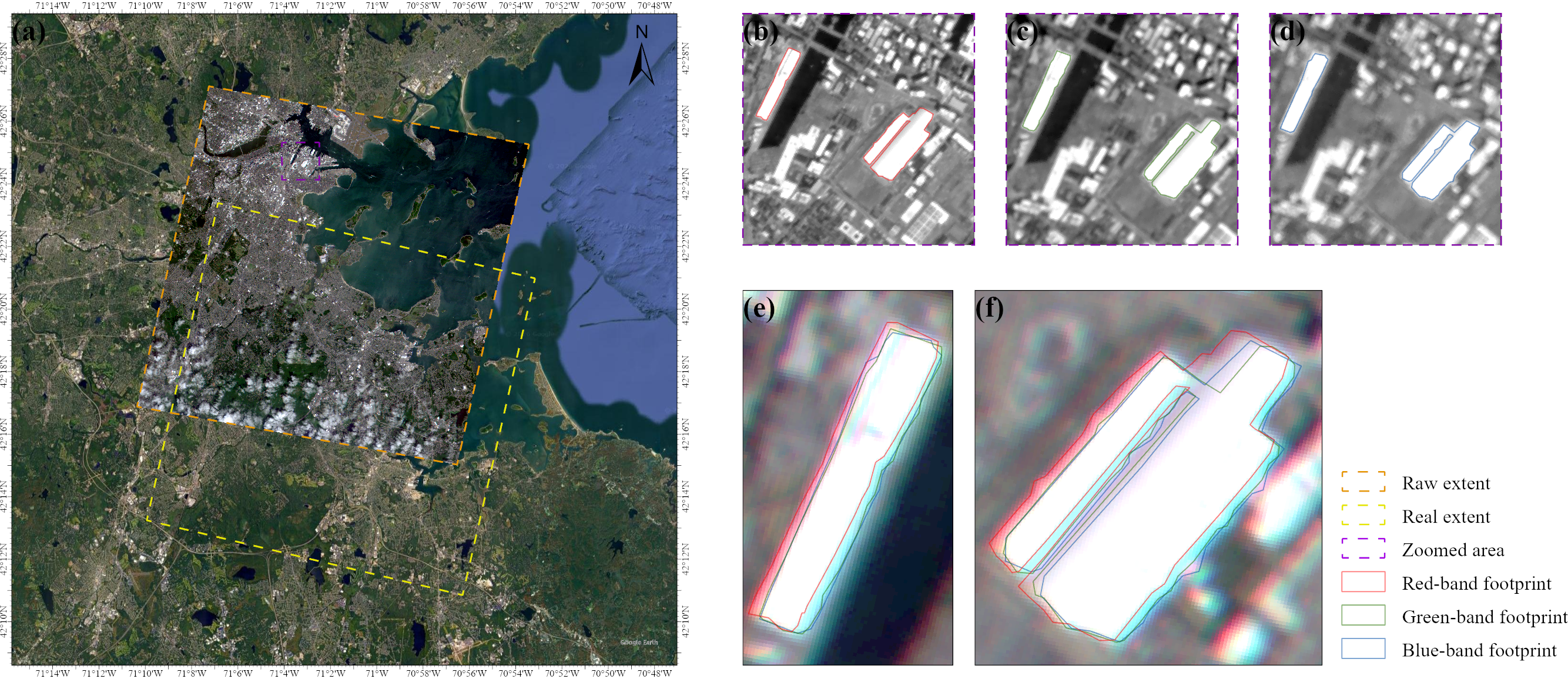}
\caption{\rev{Boston scene and local building details from PhiSat-2 imagery. (a) Scene overview with the nominal extent (orange), actual coverage (yellow), and selected detail area (purple). (b--d) Red, green, and blue bands of the selected area, respectively, with corresponding building outlines. (e,f) Enlarged RGB views of the two buildings, with band-specific outlines overlaid in matching colors.}}
\label{fig:data_quality}
\end{figure}

\rev{The Boston scene in \figref{fig:data_quality} was selected as an example with relatively pronounced geometric inconsistencies. Panel (a) shows a substantial offset between the nominal and actual scene coverage, indicating that considerable registration error remains in this officially distributed product. Panels (b--d) reveal inconsistent geometric distortions across bands, as the same building structures differ in shape and position. The RGB overlays in panels (e,f) further highlight these local discrepancies through non-coincident band-specific outlines and visible color fringing.} These phenomena are consistent with the push-broom imaging mode of PhiSat-2, where different spectral bands are acquired separately along track and may therefore retain band-to-band offsets in practical products, as noted in \hyperref[sec:phisat_2]{\textcolor{RevisionText}{Section~\ref*{sec:phisat_2}}}.

To assess the influence of such inconsistencies, a controlled inter-band shift experiment was conducted on the test set. \rev{The main TSONet checkpoints were evaluated without retraining on perturbed test images, with selected bands shifted independently in one of eight neighboring directions while the height references remained unchanged.} Two settings were considered: shifting only the non-RGB bands and shifting all spectral bands independently. \rev{Offsets were at most one pixel per axis on the 4.75\,m grid; diagonal shifts had a larger total displacement.} Pixels newly exposed by the shift operation were excluded from metric computation to avoid artificial boundary effects. \rev{The shifts were fixed across trained models using perturbation seed 2026, and all conditions were evaluated on identical valid support.}

\begin{table}[t]
\centering\tableformat
\caption{\rev{Band-offset sensitivity of frozen TSONet checkpoints on identical support.}}
\label{tab:band_offset}
\begin{tabular}{@{}lccccc@{}}
\toprule
\rev{Condition} & \rev{$\mathrm{MAE}_b\downarrow$} & \rev{$\mathrm{RMSE}_b\downarrow$} & \rev{$\mathrm{RMSE}_{\mathrm{all}}\downarrow$} & IoU$\uparrow$ & Recall$\uparrow$ \\
\midrule
\rev{Unshifted} & \rev{\metriccell{6.555}{0.108}} & \rev{\metriccell{9.921}{0.167}} & \rev{\metriccell{4.974}{0.109}} & \rev{\metriccell{34.684}{0.910}}  & \rev{\metriccell{84.266}{2.203}} \\[2pt]
\rev{Non-RGB Bands shifted} & \rev{\metriccell{6.570}{0.103}} & \rev{\metriccell{9.942}{0.155}} & \rev{\metriccell{5.019}{0.116}} & \rev{\metriccell{34.187}{0.833}}  & \rev{\metriccell{84.486}{2.225}} \\[2pt]
\rev{All Bands shifted} & \rev{\metriccell{6.548}{0.100}} & \rev{\metriccell{9.908}{0.157}} & \rev{\metriccell{5.101}{0.137}} & \rev{\metriccell{33.352}{0.719}}  & \rev{\metriccell{85.016}{1.950}} \\[2pt]
\bottomrule
\end{tabular}
\end{table}

\rev{The results in \tabref{tab:band_offset} show that band-wise shifts primarily affect building-region delineation and full-area height accuracy.} \rev{Shifting only the non-RGB bands reduces height-derived IoU from 34.684\% to 34.187\% and increases full-valid RMSE from 4.974 to 5.019\,m; shifting all bands changes these values to 33.352\% and 5.101\,m, respectively.} \rev{Building RMSE changes little, from 9.921\,m to 9.942 and 9.908\,m, while Recall increases slightly under both perturbations.} \rev{Thus, the shifts alter the spatial extent of the estimated building regions more clearly than building-height accuracy.} These results indicate that the multispectral value of PhiSat-2 depends not only on the availability of additional spectral bands, but also on their spatial co-registration quality.

These findings refine the interpretation of the sensor-information assessment in Section~\ref{sec:sensor_info}. The main experiments, conducted on real PhiSat-2 products after manual georeferencing and quality screening, indicate that current imagery already provides useful spatial--spectral information for monocular BHE. \rev{The shift experiment identifies building-region delineation as particularly sensitive to geometric consistency.} \rev{Improved band co-registration, automated geometric correction, and stricter quality screening are therefore priorities for more reliable PhiSat-2-based height estimation and large-area urban mapping.}
\rev{Consistent availability of suitable scenes and more automated preprocessing will support extension of this workflow to larger areas.}

\Needspace{5\baselineskip}

\section{Conclusion}
\label{sec:conclusion}

This study provides an initial systematic assessment of PhiSat-2 imagery for monocular building height estimation under an open-reference setting. To support this assessment, PHDataset was constructed by integrating global seven-band PhiSat-2 imagery with open building-height reference labels, resulting in 9,475 co-registered patch pairs collected from diverse urban scenes worldwide. \rev{TSONet was developed to jointly model continuous height and building-region structure.} \rev{Adaptive ordered-bin generation and pixel-wise spatial assignment are decoupled through MDHR.} \rev{Footprint guidance is incorporated at feature and reconstruction levels through FCHR.}

\rev{On the city-disjoint test subset, TSONet achieves a building RMSE of 9.928\,m.} \rev{Relative to the MHE comparison method with the lowest building-height errors, TSONet reduces RMSE and MAE by 3.64\% and 4.49\%, respectively.} \rev{Height-derived footprint delineation remains comparable, with IoU values of 34.64\% for TSONet and 36.41\% for this method.} \rev{The complete design also achieves the lowest mean building-height errors and highest height-derived IoU among the ablation configurations, supporting the combined use of MDHR and FCHR.}

\rev{The evaluated PhiSat-2 configuration reduces building RMSE by 3.73\% relative to Sentinel-2 and 6.05\% relative to Landsat 8.} \rev{The seven-band input further lowers building RMSE by 4.76\% and 4.10\% relative to RGB and RGB + NIR inputs, respectively, demonstrating the value of the fuller spectral information for height estimation.} \rev{Selected patch comparisons show more detailed urban height patterns than coarser global products, providing qualitative evidence of the potential for local urban mapping.}

\rev{In the dedicated supervision experiment, PHDataset-style open-reference supervision yields a building RMSE of 6.751\,m against LiDAR in held-out London under LiDAR-based model selection.} \rev{The separate frozen-model assessment in the main training and validation cities also shows lower building-height errors than the same MHE comparison method under an alternative reference.} \rev{Together, these experiments assess the spatial--spectral value of PhiSat-2 imagery and the usefulness of open-reference supervision for urban height mapping.}

\rev{The assessment also identifies persistent tall-building underestimation, variation across cities and reference sources, and sensitivity of footprint delineation to band misalignment. Height and footprint metrics exhibit trade-offs across methods. These findings motivate better height references, broader geographic coverage, and more consistent multispectral preprocessing.}

\section*{Acknowledgements}
This research is supported by the National Key Research and Development Program of China with grant number 2023YFB3906102, Key R \& D Projects in Yunnan Province (202403ZC38000), and Fundamental Research Fund Program of LIESMARS (4201-420100071). The numerical calculations in this paper have been done on the supercomputing system in the Supercomputing Center of Wuhan University.

\Needspace{3\baselineskip}
\section*{Declaration of generative AI and AI-assisted technologies in the manuscript preparation process}
During the preparation of this work, the authors used ChatGPT for language polishing. The authors reviewed and edited the output as needed and take full responsibility for the content of the published article. \par

\end{document}